\documentclass{article}

\usepackage[
backend=biber,
natbib=true,
style=nature, 
sorting=none
]{biblatex}

\usepackage{amsmath,amsfonts,stmaryrd,amssymb} 

\usepackage{enumerate} 

\usepackage[ruled]{algorithm2e} 

\usepackage[framemethod=tikz]{mdframed} 

\usepackage{listings} 
\usepackage{xurl} 

\usepackage{lineno} 

\usepackage{geometry} 

\usepackage[utf8]{inputenc} 
\usepackage[T1]{fontenc} 

\usepackage{XCharter} 

\RequirePackage{authblk}
\RequirePackage[labelfont={bf,sf},%
                labelsep=period,%
                justification=raggedright]{caption}

\RequirePackage[colorlinks=true, allcolors=blue]{hyperref}
\usepackage{color,soul} 

\mdfdefinestyle{commandline}{
	leftmargin=10pt,
	rightmargin=10pt,
	innerleftmargin=15pt,
	middlelinecolor=black!50!white,
	middlelinewidth=2pt,
	frametitlerule=false,
	backgroundcolor=black!5!white,
	frametitle={Command Line},
	frametitlefont={\normalfont\sffamily\color{white}\hspace{-1em}},
	frametitlebackgroundcolor=black!50!white,
	nobreak,
}

\mdfdefinestyle{file}{
	innertopmargin=1.6\baselineskip,
	innerbottommargin=0.8\baselineskip,
	topline=false, bottomline=false,
	leftline=false, rightline=false,
	leftmargin=2cm,
	rightmargin=2cm,
	singleextra={%
		\draw[fill=black!10!white](P)++(0,-1.2em)rectangle(P-|O);
		\node[anchor=north west]
		at(P-|O){\ttfamily\mdfilename};
		\def\l{3em}
		\draw(O-|P)++(-\l,0)--++(\l,\l)--(P)--(P-|O)--(O)--cycle;
		\draw(O-|P)++(-\l,0)--++(0,\l)--++(\l,0);
	},
	nobreak,
}

\mdfdefinestyle{question}{
	innertopmargin=1.2\baselineskip,
	innerbottommargin=0.8\baselineskip,
	roundcorner=5pt,
	nobreak,
	singleextra={%
		\draw(P-|O)node[xshift=1em,anchor=west,fill=white,draw,rounded corners=5pt]{%
		Question \theQuestion\questionTitle};
	},
}

\newcounter{Question} 

\mdfdefinestyle{warning}{
	topline=false, bottomline=false,
	leftline=false, rightline=false,
	nobreak,
	singleextra={%
		\draw(P-|O)++(-0.5em,0)node(tmp1){};
		\draw(P-|O)++(0.5em,0)node(tmp2){};
		\fill[black,rotate around={45:(P-|O)}](tmp1)rectangle(tmp2);
		\node at(P-|O){\color{white}\scriptsize\bf !};
		\draw[very thick](P-|O)++(0,-1em)--(O);
	}
}

\mdfdefinestyle{info}{%
	topline=false, bottomline=false,
	leftline=false, rightline=false,
	nobreak,
	singleextra={%
		\fill[black](P-|O)circle[radius=0.4em];
		\node at(P-|O){\color{white}\scriptsize\bf i};
		\draw[very thick](P-|O)++(0,-0.8em)--(O);
	}
}

\usepackage{lineno}
\usepackage{soul}
\usepackage{graphicx}
\usepackage{multirow}
\usepackage{tabularx}
\usepackage{siunitx}
\usepackage{cellspace} 
\title{\textbf{Automatic feature selection and weighting in molecular systems using Differentiable Information Imbalance}}

\author[1]{Romina Wild$^{\dag}$}
\author[2]{Felix Wodaczek$^{\dag}$}
\author[1]{Vittorio Del Tatto$^{\dag}$}
\author[3,2]{Bingqing Cheng}
\author[1,4]{Alessandro Laio*}
\affil[1]{International School for Advanced Studies (SISSA), Via Bonomea 265, Trieste, Italy}
\affil[2]{The Institute of Science and Technology Austria (ISTA), Am Campus 1, 3400 Klosterneuburg, Austria}
\affil[3]{Department of Chemistry, University of California, Berkeley, California 94720, United States}
\affil[4]{The Abdus Salam International Centre for Theoretical Physics (ICTP), Strada Costiera 11, Trieste, Italy}

\affil[$^{\dag}$]{These authors contributed equally to this work.}
\affil[*]{laio@sissa.it}

\date{November 2024}

\begin{document}
\maketitle


\section*{Abstract}

Feature selection is essential in the analysis of molecular systems and many other fields, but several uncertainties remain: What is the optimal number of features for a simplified, interpretable model that retains essential information? How should features with different units be aligned, and how should their relative importance be weighted? 
Here, we introduce the Differentiable Information Imbalance (DII), an automated method to rank information content between sets of features. Using distances in a ground truth feature space, DII identifies a low-dimensional subset of features that best preserves these relationships.
Each feature is scaled by a weight, which is optimized by minimizing the DII through gradient descent. This allows simultaneously performing unit alignment and relative importance scaling, while preserving interpretability. 
DII can also produce sparse solutions and determine the optimal size of the reduced feature space. 
We demonstrate the usefulness of this approach on two benchmark molecular problems: (1) identifying collective variables that describe conformations of a biomolecule, and (2) selecting features for training a machine-learning force field. These results show the potential of DII in addressing feature selection challenges and optimizing dimensionality in various applications. The method is available in the Python library DADApy.

\section*{Introduction}

Data sets are growing in number, in width, and in length. This abundance in data is generally used for two purposes: Predicting and understanding; likewise, feature selection has two essential aims: Model improvement and interpretability. Very often, most of the features defining a data point are redundant, irrelevant, or affected by large noise, and have to be discarded or combined, yet not many user-friendly, reliable feature selection packages exist.
For predictive modeling, feature selection is an important preprocessing step, as it helps to prevent overfitting and increases performance and efficiency \cite{Guyon2003}. In a study on leukemia cancer, for example, it was demonstrated that the disease can be best identified using just 19 out of more than 7000 genes \cite{Sarder2020}.
The other aim of feature selection is finding interpretable low-dimensional representations of high-dimensional or complex feature spaces \cite{Guyon2003}, such as those generated by molecular dynamics (MD) simulations, or learned by neural networks \cite{Wang2016}, UMAP \cite{McInnes2018, Ehiro2023} or stochastic neighbor embedding methods \cite{Maaten2008}. For example, MD trajectories produce an enormous number of variables, yet {within one graph} one can only visualize the free energy landscape in two or three dimensions that are preferably interpretable \cite{Bussi2019}. In fields like finance and medicine, finding a small number of interpretable variables is especially important for understanding the mechanisms of stock markets \cite{Yun2023} or diseases \cite{Chen2022, Remeseiro2019, Sozio2023} and can improve predictions \cite{Pathan2022}. 

Feature selection methods can be broadly divided into wrapper, embedded, and filter methods \cite{Chandrashekar2014}. 
Wrapper methods use a downstream task, such as a prediction, as the feature selection criterion, but suffer from combinatorial explosion problems. If the downstream task is akin to a classification problem, then embedded methods can perform well because they incorporate feature subset selection into the training \cite{Chandrashekar2014}. 
These algorithms are often based on regression \cite{Wu2021, Hastie1990} or on support vector machines \cite{Maldonado2011, Maldonado2018}.
Filter methods, on the other hand, are independent of a downstream task and make use of a separate criterion to rank features. They are chosen if the downstream tasks cannot be modeled easily or involve several different models.
While most wrapper and embedded methods are supervised by definition, filter methods include both, supervised and unsupervised formulations. 
Instead of using target data, unsupervised filter techniques exploit the topology of the original data manifold in various ways \cite{Liu2020, Wang2015, He2005, Cai2010, Boutsidis2009}. 
The classic supervised filters include correlation coefficient scores, mutual information \cite{Kraskov2004}, chi-square tests, and ANOVA methods \cite{Sthle1989}, which are efficient but typically consider one feature at a time, resulting in selected subsets with redundant information \cite{Guyon2003}.
Specific supervised feature subset evaluation filters like FOCUS rely on enumerating all possible subsets \cite{Almuallim1991, Urbanowicz2018}, similarly to wrapper methods, and they are affected by the same combinatorial problems.
The relief algorithm and its variants \cite{Kira1992,Urbanowicz2018} are more efficient as they do not explicitly evaluate the feature subsets. Instead, they employ nearest neighbor information to weight features, but the identified subsets can still include redundant features \cite{Urbanowicz2018}.
A review of feature selection filter methods can be found in \cite{Hopf2021}. Overall, the field of feature selection is clearly lacking the numerous powerful and out-of-the-box tools that are available in related fields such as dimensionality reduction.
 
The first, shared challenge in most of these feature selection approaches is related to the choice of the number of variables that are actually necessary to describe the system. A lower bound to such a number is provided by the intrinsic dimension \cite{Campadelli2015}, which is the dimension of the manifold containing the data. However, this number is often scale-dependent \cite{Facco2017} and position-dependent \cite{Allegra2020}. Moreover, if one wants to visualize the data {within a single graph}, the number of variables is necessarily limited to two or three. {One could show several low-dimensional projections of a high-dimensional distribution,  but this comes at the cost of readability, and a single plot is often preferable.} This typically implies neglecting part of the information, and poses the problem of choosing which variables should be retained for visualization. 

A second complication arises when the variables are heterogeneous; in many cases, a data point is defined by features with different nature and units of measures, sometimes referred to as multi-view features \cite{Zhang2019}. For example, in atomistic simulations, one can describe a molecule in water solution by providing the value of all the distances between the atoms of the molecule, which are measured in nanometers, together with the number of hydrogen bonds that they form with the solvent, which are dimensionless. In the clinical context, the features associated with a patient may include blood exams, gene expression data, and many others \cite{Wild2024}.
In order to mix heterogeneous variables in a low-dimensional description, feature selection algorithms should enable the automatic learning of feature-specific weights to correct for units of measure \cite{Zhang2019} and information content \cite{Nie2016}.

In this work, we propose a feature selection filter algorithm which mitigates many of the aforementioned problems. 
Our approach aims to find a small subset of features that can best reproduce the neighbors of the data points based on a target feature space that is assumed to be fully informative. The algorithm finds, for each input feature, an optimal \emph{weight} that accounts for different units of measure and different importance of the features. It also provides information on the optimal number of features.

The approach builds on a measure called Information Imbalance ($\Delta$), which allows comparing the information content of distances in two feature spaces \cite{Glielmo2022}.
Informally, the Information Imbalance quantifies how well pairwise distances in the first space allow for predicting pairwise distances in the second space, in terms of a score between 0 (optimal prediction) and 1 (random prediction).
This measure has been applied to find the most informative mix of containment measures for the COVID-19 pandemic \cite{Glielmo2022}, compare the information content of different machine learning interatomic potentials (MLIPs) \cite{Kandy2023}, assess the information content of chemical order parameters \cite{Donkor2023}, measure the relative information content of Smooth Overlap of Atomic Orbitals (SOAP) descriptors \cite{Darby2023}, and recently, to infer the presence of causal links in high-dimensional time series \cite{deltatto2024robust}.
In all these works, the distance space maximizing the prediction quality has been constructed by means of strategies including full combinatorial search of the optimal features \cite{Donkor2023}, greedy search approaches \cite{Wild2024} and grid search optimization of scaling parameters \cite{deltatto2024robust}, with drawbacks related to the algorithm efficiency.

Here we make a major step forward by introducing the Differentiable Information Imbalance $DII$, which allows learning the most predictive feature weights by using gradient-based optimization techniques. The input feature space, as well as the ground truth feature space (targets, labels), can have any number of features.
This provides a data analysis framework for feature selection where the optimal features and their weights are identified automatically. Moreover, carrying out the optimization with a sparsity constraint, such as L$_1$ regularization, allows finding representations of a data set formed by a small set of interpretable features.
If the full input feature set is used as ground truth, then the approach can be used as an unsupervised feature selector, whereas it acts in a supervised fashion if a separate ground truth is employed. 
To our knowledge, there is no other feature selection filter algorithm implemented in any available software package which has above mentioned capabilities. The $DII$ algorithm is publicly available in the Python package DADApy \cite{dadapy2022} and a comprehensive description can be found in the according documentation \cite{Readthedocs}, which includes a dedicated tutorial.

In the following, we will first show the effectiveness of our method on artificial examples in which the optimal set of features is known. Then we move to a real-world application and show that our approach allows addressing one of the most important challenges in molecular modeling and solid state physics: Identifying the optimal set of collective variables (CVs) for describing the configuration space of a molecular system.
As a second application, we use our method to select a subset of Atom Centered Symmetry Functions (ACSFs), descriptors of atomic environments, as input for a Behler-Parrinello machine learning potential \cite{Behler2007}, which learns energies and forces in systems of liquid water.
In the same application, we show that Smooth Overlap of Atomic Orbitals (SOAP) \cite{Bartok2013, Sandip2016} descriptors can be used as ground truth to choose informative subsets of ACSF descriptors.


\section*{Differentiable Information Imbalance} \label{sec:diff_imb}

Given a data set where each point $i$ can be expressed in terms of two feature vectors, $\mathbf{X}_i^A \in \mathbb{R}^{D_A}$ and $\mathbf{X}_i^B \in \mathbb{R}^{D_B}$ ($i=1,…,N$), the standard Information Imbalance $\Delta (d^A\rightarrow d^B)$ provides a measure of the prediction power which a distance built with features $A$ carries about a distance built with features $B$.
The Information Imbalance is proportional to the average distance rank according to $d^B$, restricted to the nearest neighbors according to $d^A$ \cite{Glielmo2022}:
\begin{equation}
\Delta\left(d^A\rightarrow d^B\right) := \frac{2}{N^2}\,\sum_{i,j:\,\, r_{ij}^A=1}  r_{ij}^B .
    \label{eq:imb}
\end{equation}
Here, $r^A_{ij}$ (resp. $r^B_{ij}$) is the distance rank of data point $j$ with respect to data point $i$ according to the distance metric $d^A$ (resp. $d^B$). For example, $r^A_{ij}=7$ if $j$ is the 7th neighbor of $i$ according to $d^A$.
$\Delta\left(d^A\rightarrow d^B\right)$ will be close to 0 if $d^A$ is a good predictor of $d^B$, since the nearest neighbors according to $d^A$ will be among the nearest neighbors according to $d^B$.
If $d^A$ provides no information about $d^B$, instead, the ranks $r_{ij}^B$ in Eq.~(\ref{eq:imb}) will be uniformly distributed between 1 and $N-1$, and $\Delta\left(d^A\rightarrow d^B\right)$ will be close to 1. {As shown in ref. }\cite{deltatto2024robust}{, the estimation of Eq. (}\ref{eq:imb}{) can potentially be improved by considering $k$ neighbors for each point.}
Considering $d^B$ as the ground truth distance, the goal is identifying the best features in space $A$ to minimize $\Delta (d^A\rightarrow d^B)$.
If the features in $A$ and the distances $d^A$ are chosen in such a way that they depend on a set of variational parameters $\mathbf{\boldsymbol{w}}$, finding the optimal feature space $A$ requires optimizing $\Delta\left(d^A(\mathbf{\boldsymbol{w}})\rightarrow d^B\right) $ with respect to $\boldsymbol{w}$.
However, $\Delta$ is defined as a conditional average of ranks, which cannot be minimized by standard gradient-based techniques.

Here we extend Eq. (\ref{eq:imb}) to a differentiable version that we call Differentiable Information Imbalance ($DII$) in order to automatically learn the optimal distance $d^A(\boldsymbol{w})$.
We approximate the non-differentiable, rank-dependent sum in Eq.~(\ref{eq:imb}) by introducing the softmax coefficients $c_{ij}$:
\begin{equation}\label{eq:diff_II}
    DII\left(d^A(\boldsymbol{w})\rightarrow d^B\right) := \frac{2}{N^2}\,\sum_{\substack{i,j=1 \\ (j\neq i)}}^N c_{ij}(\lambda,d^A(\boldsymbol{w}))\,r_{ij}^B  \,,
\end{equation}
where
\begin{equation} \label{eq:softmax_coeffs}
    c_{ij}(\lambda,d^A(\boldsymbol{w})) := \frac{e^{-d^A_{ij} (\boldsymbol{w})/\lambda}}{\sum_{m (\neq i)} e^{-d^A_{im}( \boldsymbol{w})/\lambda} } .
\end{equation}
The coefficients $c_{ij}$ in Eq.~(\ref{eq:diff_II}) approximate the constraint $r_{ij}^A = 1$, such that $ c_{ij} \rightarrow \delta_{1,r_{ij}^A}$ as $\lambda\rightarrow 0$ ($\delta$ denotes the Kronecker delta). Therefore, as illustrated in the tutorial \texttt{Differentiable Information Imbalance} in ref. \cite{Readthedocs}, in the limit of small $\lambda$ the $DII$ converges to $\Delta$:
\begin{equation}\label{eq:diff_equiv}
    \lim_{\lambda\rightarrow 0} DII\left(d^A(\boldsymbol{w})\rightarrow d^B\right) = \Delta\left(d^A(\boldsymbol{w})\rightarrow d^B\right)\,.
\end{equation}
For any positive and small $\lambda$, the quantity $DII\left(d^A\rightarrow d^B\right)$ can be seen as a continuous version of the Information Imbalance, where the coefficients $c_{ij}$ assign, for each point $i$, a non-zero and exponentially decaying weight to points $j$ ranked after the nearest neighbor in space $d^A$. The parameter $\lambda$ is chosen according to the average and minimum nearest neighbor distances (see \nameref{sec:adalambda}).

The $DII$ is differentiable with respect to the parameters $\boldsymbol{w}$ for any distance $d^A$ which is a differentiable function of $\boldsymbol{w}$. 
In this work, we assume that the variational parameters are weights, $\boldsymbol{w} = (w^1, ...\,, w^{D_A})$, scaling the features in space $A$ as $\boldsymbol{w}\odot\mathbf{X}_i^A = (w^1\,X_i^1, ...\,, w^{D_A}\,X_i^{D_A})\,$ (the symbol $\odot$ denotes the element-wise product).
We construct $d^A(\boldsymbol{w})$ as the Euclidean distance between these scaled data points, $d^A_{ij} (\boldsymbol{w})=\|\boldsymbol{w}\odot\left(\mathbf{X}^A_i - \mathbf{X}^A_j\right) \|$.
In this case, the coefficients $c_{ij}$ can be written as
\begin{equation} \label{eq:softmax_coeffs2}
    c_{ij}= \frac{e^{- \| \boldsymbol{w}\odot\left(\mathbf{X}_i^{A} - \mathbf{X}_j^{A}\right) \|/\lambda}}{\sum_{m(\neq i)} e^{- \|\boldsymbol{w}\odot\left(\mathbf{X}_i^{A} - \mathbf{X}_m^{A} \right) \|/\lambda}} ,
\end{equation}
and the derivatives of $DII\left(d^A(\boldsymbol{w})\rightarrow d^B\right)$ with respect to the parameters $w^\alpha$ can be computed:
\begin{equation}\label{eq:derivative}
    \frac{\partial}{\partial w^\alpha} DII\left(d^A (\boldsymbol{w})\rightarrow d^B\right) = \frac{2\,w^\alpha}{\lambda\, N^2} \sum_{\substack{i,j \\ (i\neq j)}} c_{ij}\,r_{ij}^B \left(-\frac{(X_i^\alpha -X_j^\alpha)^2}{\| \boldsymbol{w} \odot \left(\mathbf{X}_i^{A} - \mathbf{X}_j^{A} \right) \|} + 
    \sum_{m (\neq i)} c_{im} \frac{(X_i^\alpha -X_m^\alpha)^2}{\| \boldsymbol{w} \odot \left( \mathbf{X}_i^{A} - \mathbf{X}_m^{A}\right) \|}\right)\,.
\end{equation}
These derivatives can be used in gradient-based methods to minimize the $DII$ with respect to the variational weights.

If one aims at a low-dimensional representation of the feature space $A$, as in the case of feature selection, it is desirable that several of the weights are set to zero.
While for up to $D_A\sim 10$ a full combinatorial search of all feature subsets can be carried out, optimizing the $DII$ over each subset, for larger feature spaces a sparsification heuristic becomes necessary.
We complement the $DII$ optimization with two approaches for learning sparse features: Greedy backward selection and L$_1$ (lasso) regularization.
Greedy selection removes one feature at a time from the full set, according to the lowest weight. L$_1$ regularization selects the subset of features that optimizes the $DII$ while simultaneously keeping the L$_1$ norm of the weights small (see \nameref{sec:l1}).
While greedy backward selection gives reliable results for up to $\approx$ 100 features, in larger feature spaces this algorithm becomes computationally demanding, and it is advisable to use L$_1$ regularization to find sparse solutions.

\section*{Results}

\subsection*{Benchmarking the approach: Gaussian random variables and their monomials}

We first test the $DII$ approach using two illustrative examples where the distances $d^A(\boldsymbol{w})$ and $d^B$ are built with the same features, so that the target weights minimizing Eq. (\ref{eq:diff_II}) are known.
In particular, we take as ground truth distance $d^B$ the Euclidean distance in the space of the scaled data points $\boldsymbol{w}_{GT}\odot\mathbf{X}_i$, where the weights $\boldsymbol{w}_{GT}$ are fixed and known.
We aim at recovering the target weights by scaling the unscaled input features, $\boldsymbol{w}\odot\mathbf{X}_i$, with the proposed $DII$-minimization.

In each example, we carry out several optimizations, both without any regularization term and in presence of a L$_1$ penalty, which induces sparsity in the learned weights.
For each optimization, we employ a standard gradient descent algorithm, initializing the parameters $\boldsymbol{w}$ with the inverse of the features' standard deviations (see \nameref{sec:opti} for further details).
In order to judge the quality of the recovered weights in the various settings, we calculate the cosine similarity between the vector of the optimized weights and $\boldsymbol{w}_{GT}$.
This evaluation metric, which is bounded between 0 (minimum overlap) and 1 (maximum overlap), only depends on the relative angle between the two vectors, reflecting the fact that the $DII$ allows to recover the target weights up to a uniform scaling factor (\nameref{sec:invar}). 

\begin{figure}[!ht]
    \centering
    \includegraphics[width=.89\textwidth]{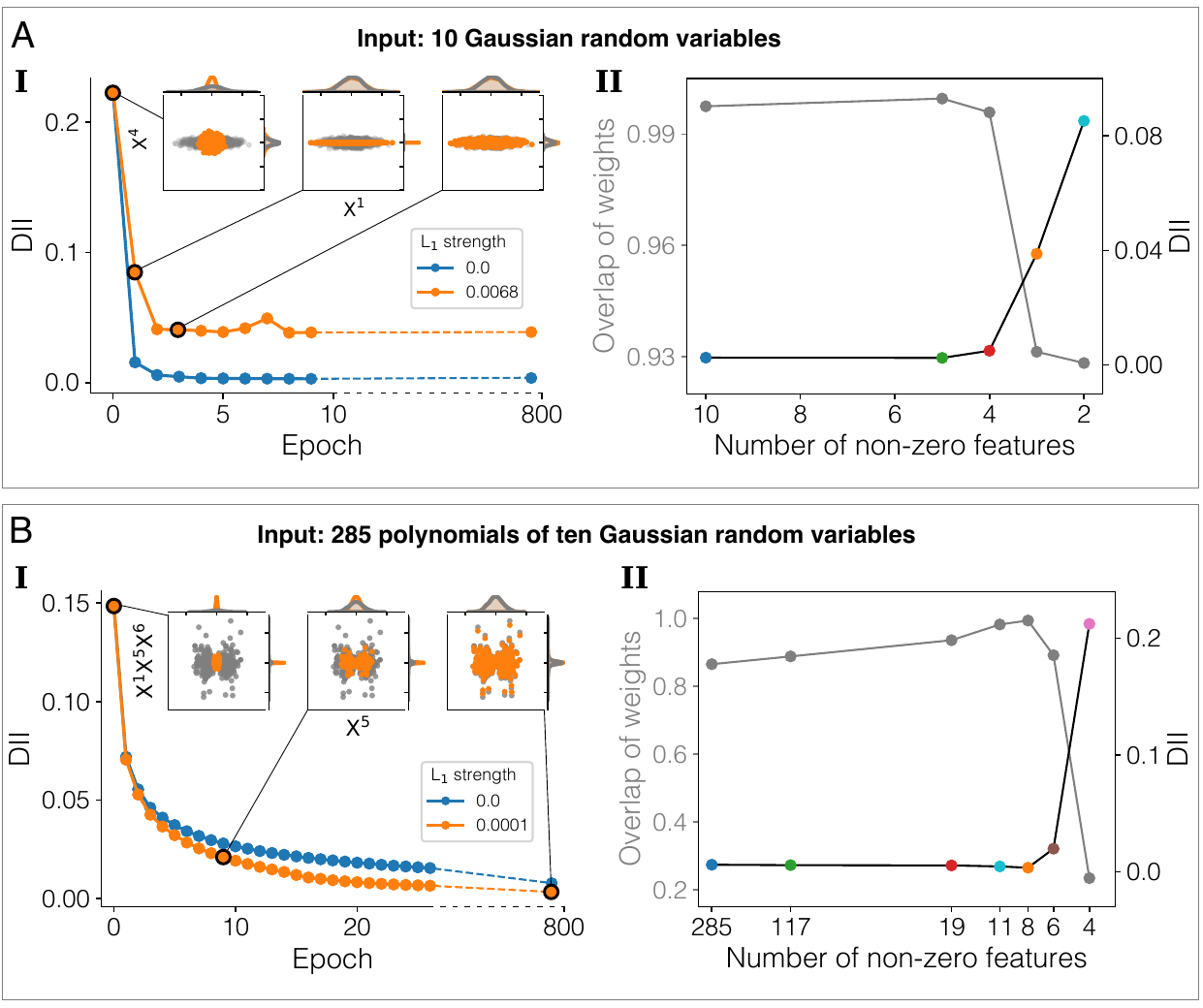}
    \caption{\textbf{$\boldsymbol{DII}$ feature selection applied to Gaussian random variables and their monomials. A}: The input features are ten independent and identically distributed Gaussian random variables, X$^1$-X$^{10}$. The same features are used as ground truth, but scaled. \textbf{I}: Differentiable Information Imbalance ($DII$), with (orange) and without (blue) L$_1$ regularization in the optimization. The insets show two exemplary features, with the weights during optimization (orange) and the ground truth weights (gray). \textbf{II}: Cosine similarity (overlap) of the ground truth and optimized weights in gray, and $DII$s in black with colored markers, for several L$_1$ strengths and associated numbers of non-zero features. Table \ref{table:1} provides the ground truth and optimized weights for points in this graph. \textbf{B}: The feature space consists of the 285 monomials up to order three of the ten Gaussian random variables from \textbf{A}. As ground truth, ten features were selected at random and scaled, while all the other feature weights are zero. \textbf{I} and \textbf{II}: Analogous to \textbf{A}. Table \ref{table:2} provides the ground truth and optimized weights for points in this graph.
    Source data are provided as a Source Data file.}
    \label{fig:FIG1}
\end{figure}

\begin{table}[!ht]
\caption{\textbf{Ground truth weights, optimized weights and optimization details for the 10 Gaussian random variables corresponding to Fig. \ref{fig:FIG1}A}: The feature space consists of ten independent and identically distributed Gaussian random variables, X$^1$-X$^{10}$. The same features are used as ground truth, but scaled. Optimized weights are shown at selected L$_1$ regularization strengths (L$_1$ reg.), and the resulting number of nonzero features (Nnz.) and Differentiable Information Imbalance ($DII$) are provided.}
\label{table:1}
\small
\resizebox{\textwidth}{!}{%
\setlength{\extrarowheight}{2pt} 
\begin{tabularx}{\textwidth}{*{10}{>{\centering\arraybackslash}X}|*{3}{>{\centering\arraybackslash}X}}
\hline
\multicolumn{10}{c|}{\textbf{Features}} & \multirow{5}{*}{\textbf{L$_1$ reg.}} & \multirow{5}{*}{\textbf{Nnz.}} & \multirow{5}{*}{\textbf{$\mathbf{DII}$}} \\ \cline{1-10}
\textbf{$X^1$} & \textbf{$X^2$} & \textbf{$X^3$} & \textbf{$X^4$} & \textbf{$X^5$} & \textbf{$X^6$} & \textbf{$X^7$} & \textbf{$X^8$} & \textbf{$X^9$} & \textbf{$X^{10}$} &  &  &  \\ \cline{1-10}
\multicolumn{10}{c|}{\textbf{Ground truth weights}} &  &  &  \\ \cline{1-10}
5.0 & 2.0 & 1.0 & 1.0 & 0.5 & $10^{-4}$ & $10^{-4}$ & $10^{-4}$ & $10^{-4}$ & $10^{-4}$ &  &  &  \\ \cline{1-10}
\multicolumn{10}{c|}{\textbf{$\mathbf{DII}$ optimized weights}} &  &  &  \\ \hline
5.0 & 2.3 & 1.2 & 1.2 & 0.6 & $10^{-9}$ & $10^{-12}$ & $10^{-9}$ & $10^{-9}$ & $10^{-9}$ & None & 10 & 0.003 \\ \hline
5.0 & 2.1 & 1.1 & 1.1 & 0.5 & 0 & 0 & 0 & 0 & 0 & 0.0001 & 5 & 0.002 \\ \hline
5.0 & 2.0 & 1.1 & 1.1 & 0 & 0 & 0 & 0 & 0 & 0 & 0.0002 & 4 & 0.005 \\ \hline
5.0 & 0.5 & 0 & 0.2 & 0 & 0 & 0 & 0 & 0 & 0 & 0.0068 & 3 & 0.039 \\ \hline
5.0 & 0.6 & 0 & 0 & 0 & 0 & 0 & 0 & 0 & 0 & 0.01 & 2 & 0.085 \\ \hline
\end{tabularx}%
}
\end{table}

\begin{table}[!ht]
\caption{\textbf{Ground truth weights, optimized weights and optimization details for the 285 monomials corresponding to Fig. \ref{fig:FIG1}B}: The feature space consists of the 285 monomials up to order three of ten Gaussian random variables. As ground truth, ten features were selected at random and scaled, while all the other feature weights are zero. Optimized weights are shown at selected L$_1$ regularization strengths (L$_1$ reg.), and the resulting number of nonzero features (Nnz.) and Differentiable Information Imbalance ($DII$) are provided. The sum of the remaining 275 non-ground-truth weights is shown (Sum).}
\label{table:2} 
\small
\resizebox{\textwidth}{!}{%
\setlength{\extrarowheight}{2pt} 
\begin{tabularx}{\textwidth}{*{11}{>{\centering\arraybackslash}X}|>{\centering\arraybackslash}p{1.2cm}>{\centering\arraybackslash}p{0.9cm}>{\centering\arraybackslash}p{0.9cm}}

\hline
\multicolumn{11}{c|}{\textbf{Features}} & \multirow{5}{*}{\textbf{L$_1$ reg.}} & \multirow{5}{*}{\textbf{Nnz.}} & \multirow{5}{*}{\textbf{$\mathbf{DII}$}} \\ \cline{1-11}
$X^5$ & $X^1\!X^5\!X^6$ & $\quad\, X^3$ & $(X^2)^2$ & $X^6$ & $X^{10}$ & $X^1\!X^2$ & $X^8\!(X^{10})^2$ & $ \quad\,\, X^8$ & $X^5\!X^8$ & Other &  &  &  \\ \cline{1-11}
\multicolumn{10}{c}{\textbf{\hspace{1cm}Ground truth weights}} & \textbf{Sum} &  &  &  \\ \cline{1-11}
10.0 & 7.0 & 6.0 & 5.0 & 5.0 & 4.0 & 3.0 & 2.0 & 1 .0 & 1.0 & 0 &  &  &  \\ \cline{1-11}
\multicolumn{11}{c|}{\textbf{$\mathbf{DII}$ optimized weights}} &  &  &  \\ \hline
10.0 & 6.2 & 6.9 & 2.9 & 6.1 & 5.1 & 2.2 & 2.5 & 0.8 & 0.7 & 72.4 & None & 285 & 0.006 \\ \hline
10.0 & 6.2 & 6.9 & 2.9 & 6.1 & 5.1 & 2.2 & 2.4 & 0.8 & 0 & 53.0 & \num{4e-6} & 117 & 0.006 \\ \hline
10.0 & 4.5 & 6.0 & 1.8 & 5.1 & 4.0 & 1.6 & 1.6 & 0 & 0 & 8.9 & \num{5e-4} & 19 & 0.005 \\ \hline
10.0 & 6.2 & 6.5 & 3.7 & 5.6 & 4.5 & 3.1 & 2.1 & 0.7 & 0 & 2.6 & \num{5e-5} & 11 & 0.005 \\ \hline
10.0 & 6.3 & 6.3 & 5.0 & 5.4 & 4.3 & 3.0 & 2.1 & 0 & 0 & 0 & 0.0001 & 8 & 0.003 \\ \hline
10.0 & 6.9 & 6.2 & 0 & 5.7 & 3.1 & 0 & 0 & 0 & 0 & 3.8 & 0.0014 & 6 & 0.020 \\ \hline
2.3 & 1 .1 & 1.4 & 0 & 0 & 0 & 0 & 0 & 0 & 0 & 10 & 0.0038 & 4 & 0.212 \\ \hline
10.0 & 0 & 0 & 0 & 0 & 0 & 0 & 0 & 0 & 0 & 0 & 0.0023 & 1 & 0.606 \\ \hline
\end{tabularx}%
}
\end{table}

In the first example, we use a data set of 1500 points drawn from a 10-dimensional Gaussian with unit variance in each dimension, and we construct a ground truth distance $d^B$ by assigning non-zero weights $w_{GT}^\alpha$ to all its 10 components (Table \ref{table:1}).
The target weights $w_{GT}^6$ to $w_{GT}^{10}$ are close to zero, such that these features carry almost no information.

The optimization without any L$_1$ regularization yields a very good result in terms of $DII$ and overlap (blue in Fig.~\ref{fig:FIG1}A I and II). If a soft L$_1$ regularization strength is employed, the results are qualitatively the same, but the irrelevant features $\alpha=6$-$10$ receive zero weights, inducing sparsity and leading to an effective feature selection (green in Fig.~\ref{fig:FIG1}A II). Table \ref{table:1} shows the learned weights for different strengths of the L$_1$ penalty, scaled in such a way that the largest weight is identical to the largest component of $\boldsymbol{w}_{GT}$.
Since in $DII$ only the relative weights are important, this scaling is permissible and helps illustrating the comparison.
By increasing the regularization strength, more features are set to zero following the order of their ground truth weights. When features of higher importance, namely with higher ground truth weights, are forced to zero by the regularization, then the resulting $DII$ increases and the cosine similarity decreases, showcasing the loss of information (Fig.~\ref{fig:FIG1}A II, Table \ref{table:1}).

Secondly, to test the method in a high-dimensional setting, we created a data set with 285 features including all the products up to order three of the 10 Gaussian random variables used in the previous example. 
Products of Gaussian random variables are distributed according to Meijer \textit{G}-functions, which may not be Gaussian \cite{Springer1970}.
The ground truth distance $d^B$ is here built by only selecting ten of these monomials, with various weights (Table \ref{table:2}). All other feature weights in the ground truth can be considered zero. 

Since in this case the correct solution is very sparse in the full feature space, an appropriate sparsity-inducing regularization becomes essential to obtain good results. Without any L$_1$ regularization, all the 285 features receive a non-zero weight. 
Even if, in this case, the ground truth features are assigned the highest weights, there might not be a clear cut-off in the weight spectrum to distinguish them from the less-informative features. 

As shown in Fig.~\ref{fig:FIG1}B, the correct level of regularization can be identified by computing the $DII$ as a function of the non-zero features or regularization strength. 
The intermediate L$_1$ strength of $0.0001$ results in the best performance, as it coincides with the lowest $DII$ and the largest weight overlap (orange in Fig.~\ref{fig:FIG1}B I and II). The eight most relevant ground truth features are correctly identified, with an overlap between the learned and the ground truth weights which is remarkably close to 1.

Furthermore, panel I in Fig.~\ref{fig:FIG1}B shows that weights found with L$_1$ regularization have a lower $DII$ than the ones without L$_1$ regularization in the same optimization time, which means that the weights resulting from a certain level of regularization are effectively better than the unregulated ones.
As in the previous example, when the regularization is too strong, some of the relevant features are discarded, resulting in a drop in the weight overlap and an increase in the $DII$ (Fig.~\ref{fig:FIG1}B II, Table \ref{table:2}).

We then benchmarked the $DII$ method against other feature selection methods. We perform the benchmark on the example with 285 monomials, in which the ground truth is known. 

There are very few methods available in software packages which can be applied to the specific task we are considering, which is selecting and scaling features from a high-dimensional input space to be maximally informative about a multi-dimensional continuous ground truth,  defining a pairwise \emph{distance}. Considering filter methods, we compare $DII$ to relief-based algorithms (RBAs), specifically RReliefF and MultiSURF, implemented in scikit-rebate {\cite{Urbanowicz2018bioinformatics}}, which support a continuous ground truth {\cite{Urbanowicz2018}}. RBAs are filter methods that weight features, but importantly only work with a one-dimensional ground truth. 
This poses a problem for all use cases in this paper because the ground truth is always defined by the multi-dimensional vector of features used to compute the target distance. 
RBAs extended to the multi-label case {\cite{Spolaor2013, Zhang2023}} but, to our knowledge, are not implemented in software packages. We apply scikit-rebate RReliefF and MultiSURF for each ground truth dimension individually, and sum the resulting weights with and without prior importance cutoff (see Suppl. Inf.). The methods detect the most important input feature in most cases, leading to overall cosine similarities ranging from $0.56$ to $0.84$ for the various settings (Suppl. Fig. 1).

As a second benchmark we use a method from scikit-learn {\cite{scikit-learn}}, which can handle the task's requirements: The decision tree regressor ({\texttt{sklearn.tree.DecisionTreeRegressor}}). Unlike $DII$ and the relief-based algorithms, this method is not a filter but an embedded method. The feature selection is determined as a side product during the building of a regressor model. There is no filter algorithm implemented in scikit-learn which can solve a problem as posed here. Two distinct feature importance measures implemented with the approach, the Gini importance and the Permutation importance, lead to feature vectors with a cosine similarity of up to $0.83$ with respect to the ground truth. 
In comparison, the $DII$ method with a L$_1$regularization of $0.0001$ (orange in Fig. {\ref{fig:FIG1}B}) finds a weight vector with eight non-zero weights and a cosine similarity of $0.99$.

In conclusion, in both examples the $DII$ method is able to recover the ground truth weights with good accuracy, and better than the very few other applicable methods, as measured by a larger weight overlap with the ground truth.
In the following sections, we apply our feature selection method to cases in which the optimal solution is not known and illustrate how our approach can be used to give an explicit system description by extracting few features from a larger data set.

\subsection*{Identifying the optimal collective variables for describing a free energy landscape of a small peptide}

We now illustrate how the $DII$ can be used to identify the most informative collective variables (CVs) to describe the free energy landscape of a biomolecule. {As opposed to the previous example, in this test the ground truth variables and the input variables are different sets. }

We consider a temperature replica-exchange MD simulation (400 ns, 340 K replica analyzed only, $dt$ = 2 fs) \cite{Carli2021} of the 10-residue peptide CLN025 \cite{Honda2004}, which folds into a $\beta$-hairpin.
The data set is composed of 1429 frames (subsampled from 41,580 trajectory frames) containing all atom coordinates. 
The ground truth metric $d^B$ is constructed in the feature space of all the 4,278 pairwise distances between the 93 heavy atoms of the peptide, which can be assumed to hold the full conformational information of the system.
We consider a feature space $A$ with ten classical CVs that do not depend on knowledge of the folded state of the $\beta$-hairpin peptide: Radius of gyration (RGYR), anti-$\beta$-sheet content, $\alpha$-helical content, number of hydrophobic contacts, principal component 1 (PC1), principal component 2 (PC2), principal component residuals, the number of hydrogen bonds in the backbone, in the side chains, and between the backbone and side chains (see \nameref{sec:cvs}). 

\begin{figure}[!ht]
    \centering
    \includegraphics[width=.99\textwidth]{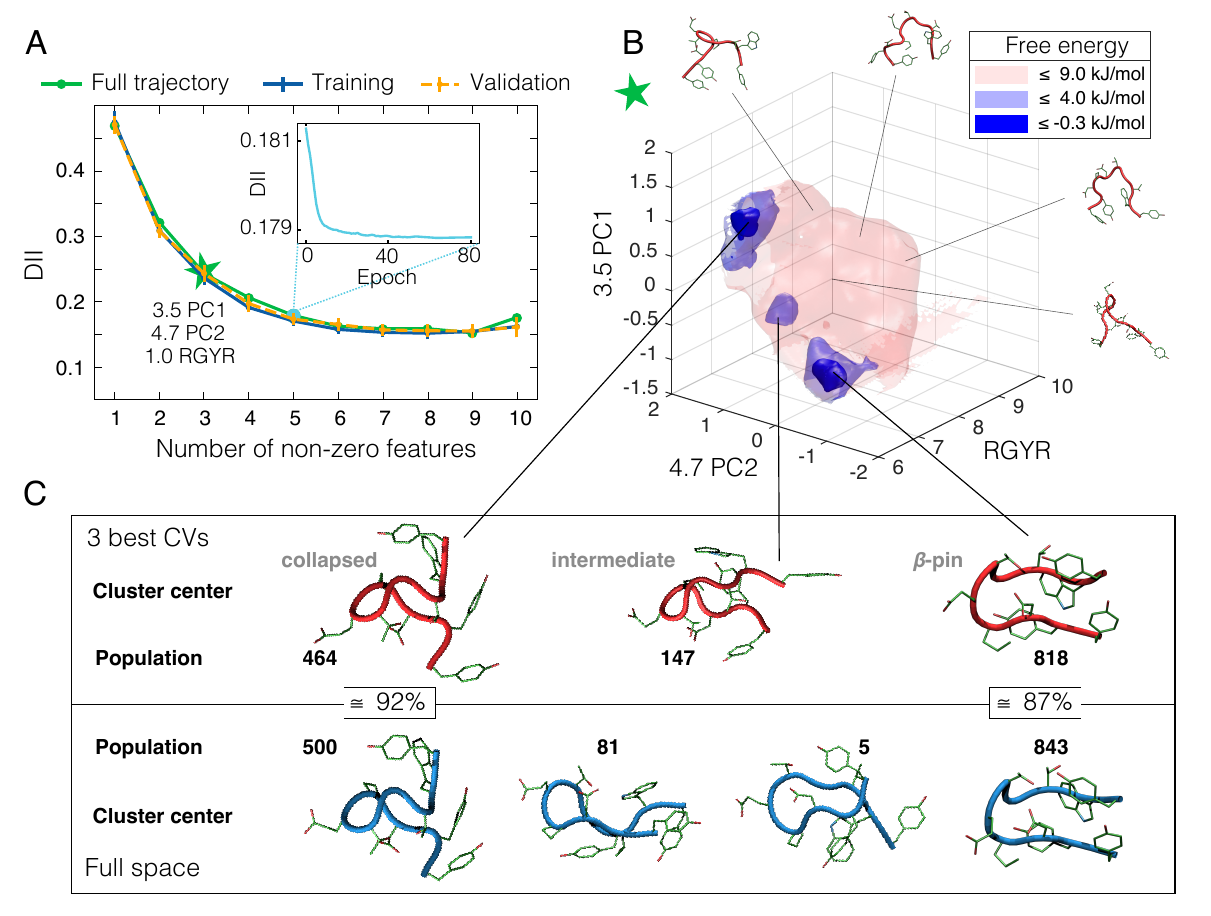}
    \caption{\textbf{$\boldsymbol{DII}$ feature selection for describing the free energy landscape and conformations of CLN025. A:} Green: Optimal Differentiable Information Imbalance ($DII$) results for collective variable (CV) subsets of different sizes with gradient descent optimized weights for 1429 data points evenly sampled from the full trajectory. The green star marks the $DII$ result of the optimally scaled 3-plet, which defines the coordinate system for \textbf{B}. Inset: $DII$ gradient descent optimization for the optimal 5-plet. Blue and orange: Average and standard deviations of the $DII$ calculated from block cross validation with 4 non-overlapping training data sets and 84 validation sets of 1428 points each. 
    \textbf{B}: Free energy isosurfaces in the space of the optimal 3-plet of CVs (radius of gyration (RGYR), principal components 1 and 2 (PC1 and PC2), with weights of 1.0, 3.5 and 4.7), corresponding to three different values of the free energy. The renderings around the free energy surfaces show sampled conformations of the peptide at different values of the CVs and free energy.
    \textbf{C}: Red and blue renderings are cluster centers obtained from the optimal 3-plet space and from the full space of all pairwise heavy atom distances, respectively. The two main cluster centers of both belong to the dominant peptide conformations: The $\beta$-pin and the collapsed denatured state. The collapsed and $\beta$-pin clusters identified in the optimal 3-plet space share 92\% and 87\% of the frames with the corresponding full space clusters.
    Source data are provided as a Source Data file.}
    \label{fig:CLN1}
\end{figure}

Since the CV feature space is only 10-dimensional, it is possible to look for the optimal distance $d^A$ by an exhaustive search of all possible 1023 subsets containing one to ten CVs, without using the L$_1$ regularization to produce sparse solutions.
For each subset of CVs, the $DII$ is used as a loss to automatically optimize the scaling weights, 
which are initialized to the inverse standard deviations of the corresponding variables.
Even when all feature subsets can be constructed, gradient descent optimization of the $DII$ is useful, as the most naive choices of the scaling weights - setting them to the inverse standard deviations of the variables, or all equal to 1 - likely define suboptimal distances, since the CVs have different units of measure and importance.
The optimization of the feature weights for all 1023 subsets takes about 4.5 h on a CentOS Linux 7 with 24 CPUs Intel Xeon E5-2690 (2.60GHz) with 15 GB RAM using the function ``return\_weights\_optimize\_dii'' with 80 epochs (Fig.~\ref{fig:CLN1}A, green curve).

Fig.~\ref{fig:CLN1}A shows the results of the subset optimizations by computing the $DII$ with block cross-validation (see \nameref{sec:crossval}).
The training and validation $DII$s averaged over all cross-validation splits, show a high degree of consistency, verifying the transferability of the $DII$ results between non-overlapping pieces of the trajectory. As shown in the inset graph in Fig.~\ref{fig:CLN1}A, the $DII$ result improves during the gradient descent optimization. 
The best single CV is anti-$\beta$-sheet content, while the best triplet contains RGYR, PC1 and PC2 with weights of $1.0$, $3.5$ and $4.7$. Remarkably, the weight of PC2 is higher than the weight of PC1, confirming that the gradient optimization of the $DII$ provides non-trivial results.
We estimated the density in the space of the best three scaled variables (Fig.~\ref{fig:CLN1}B) using point-adaptive $k$-NN (PA$k$) \cite{Rodriguez2018}, implemented in the DADApy package \cite{dadapy2022}.
The free energy derived from this density clearly shows two favorable main states, which are the folded $\beta$-hairpin state and a denatured collapsed state \cite{McKiernan2017} with negative values of the free energy in Fig.~\ref{fig:CLN1}B.

The cluster centers found by Density Peak Clustering in its unsupervised extension \cite{dErrico2021} are depicted by the renderings denoted ``collapsed'', ``intermediate'', and ``$\beta$-pin'' in Fig.~\ref{fig:CLN1}C, while additional example structures from less favorable free energy regions are shown around Fig.~\ref{fig:CLN1}B. The clustering was also performed in the full space of all 4,278 heavy atom distances, which holds the full information of the system.

The populations of both, $\beta$-pin and collapsed clusters show a remarkable overlap between the clustering structures obtained in the optimal 3-plet case and from the full feature space of 4,278 heavy atom distances.
Taking the cluster populations from the full space as ground truth classes, such overlap can be simply measured as the fraction of points (trajectory frames) that belong to the same cluster in both representations, also referred to as cluster purity \cite{Manning2008}: The $\beta$-hairpin cluster from the 3-plet space has 87\% purity, and the collapsed state cluster has 92\% purity, considering the full space as reference. Taken together, all clusters have a 89$\%$ overall cluster purity towards the full space clusters. 
This consistency also emerges by visually comparing the red and blue renderings of the two dominant cluster centers (left and right structures in Fig.~\ref{fig:CLN1}C).
As a comparison, running the clustering algorithm using the single best CV, the anti-$\beta$-sheet content, brings to an overall cluster purity of 45\%, i.e. the trajectory frames clustered into the pin, collapsed, or other clusters using the single best variable, capture 45\% of the same frames of the according clusters using the full space for clustering. Hence, no single one-dimensional CV is informative enough to describe CLN025 well, but a combination of only three scaled CVs carries enough information to achieve an accurate description of this system.

Because of the good performance of decision tree regression on the previous example and its ability to handle multi-target (even high-dimensional), continuous ground truth data, we apply this feature selection algorithm also to this use case (Suppl. Inf.). The best three variables using the Gini importance weights are: $0.29$ anti-$\beta$-sheet content, $0.25$ PC1, $0.1$ PC2; using the permutation importance they are: $1.27$ PC1, $1.04$ anti-$\beta$-sheet content, $0.97$ PC2. Clustering in these reduced spaces leads to maximum cluster purities compared to the full space clusters of 55\% for Gini importance and 63\% for the permutation importance and several additional inconsistencies when compared to the full space clustering (Suppl. Fig. S2 and suppl. text).

We also test the robustness of the method using four uncorrelated trajectory blocks and performing the $DII$-optimization in each of these blocks. The resulting $DII$s, as well as selected features and their weights, show excellent consistency  across the blocks (Suppl. Fig. S4 and S5).

\subsection*{Feature selection for Machine Learning Potentials}

In another use case of the $DII$ approach, we demonstrate its capabilities for selecting features for training Behler-Parrinello machine learning potentials (MLPs) \cite{Behler2007}.
MLPs can learn energy and forces of atomic configurations derived from quantum mechanical calculations. The Behler-Parrinello MLP uses Atom Centered Symmetry Functions (ACSF) as inputs for the predictions \cite{Behler2011}. The ACSFs are a large set of radial and angular distribution functions, which describe the environment around an atom, and are permutationally, rotationally, and translationally invariant.

\begin{figure}[!ht]
    \centering
    \includegraphics[width=0.6\textwidth]{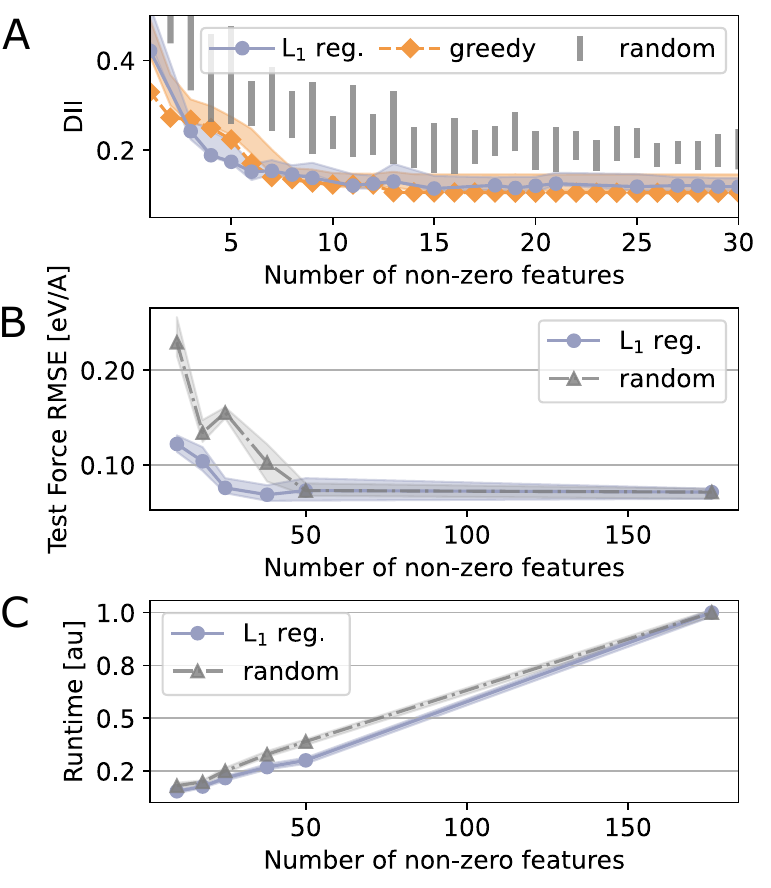}
    \caption{
        \textbf{$\boldsymbol{DII}$ feature selection for efficient training of a Machine Learning Potential (MLP). A}: Differentiable Information Imbalance ($DII$) selecting the optimal feature subsets from $D_A=176$ Atom Centered Symmetry Functions (ACSF) descriptors, against a ground truth of $D_B=546$ Smooth Overlap of Atomic Orbitals (SOAP) descriptors, using a data set of $N\sim 350$ atomic environments.
        The optimized $DII$ per number of non-zero features is shown by blue circles and orange diamonds, using L$_1$ regularized search and greedy backward selection, respectively.
        The filled area represents validation data in the form of the minimum and maximum $DII$ on 10 batches of $\sim$350 atomic environments other than the $\sim$350 environments used for $DII$ feature selection.
        The $DII$ for randomly selecting a certain number of non-zero features is depicted as gray bars between the lowest and highest $DII$ found within 10 random selections. 
        \textbf{B}: Test root-mean-square error (RMSE) with features chosen via L$_1$ regularized $DII$ (blue circles) and at random (gray triangles) by Behler-Parrinello-type MLPs \cite{Behler2007} as implemented in n2p2 \cite{Singraber2019, Singraber2019_2}.
        {Six MLPs with different train-test splits per number of non-zero features are trained.
        Markers represent their average RMSE, the filled area shows the range from worst to best performer.}
        \textbf{C}: Run-time of force and energy prediction on a single structure performed by the same MLPs as in \textbf{B}.
        The filled area shows the range from worst to best performer, despite being barely visible due to similar run-times across the six MLPs.
        Source data are provided as a Source Data file.
    }
    \label{fig:force_prediction}
\end{figure}

The data set used here consists of $N\sim350$ atomic environments of liquid water molecules, derived from a larger data set that has previously been used to fit a MLP, which can accurately predict various physical properties of water \cite{Cheng2018}. 
The input features in this example are 176 ACSF descriptors (see \nameref{sec:mlp}). The ACSF descriptor dimensions combinatorially grow with the number of atom types, which makes them computationally costly and makes feature selection attractive \cite{Keith2021}. Since the ACSF space is too large for full combinatoric feature selection, we search for sparse solutions using both L$_1$ regularized $DII$ and greedy backward selection (``L$_1$ reg.'' and ``greedy'' in Fig.~\ref{fig:force_prediction}, see \nameref{sec:l1}).
We aim to select informative ACSFs before the training to reduce the number of input features and thus reduce the training and prediction time.

While $DII$ can be used as an unsupervised feature selector when the feature space is reduced against itself as ground truth, it can also incorporate a separate feature space as ground truth in a supervised fashion. This is especially useful when a comprehensive ground truth exists. In the case of atomic environments, one of the most complete, accurate, and robust descriptions is given by the Smooth Overlap of Atomic Orbitals (SOAP) descriptors \cite{Bartok2013, Sandip2016}, based on the expansion of the local density in spherical harmonics. 546 SOAP features (n$_{\text{max}}=6$, l$_{\text{max}}=6$) are defined as the ground truth for feature selection.
In this manner, we can put SOAP and ACSF, two comprehensive representations of atomic environments, into relation \cite{Musil2021} and show that SOAP is a suitable ground truth to select informative ACSFs as inputs for a MLP. {The SOAP space captures the full spacial arrangement of atoms by encoding the local atomic densities and accounting for symmetries} \cite{Bartok2013}. Both SOAP and ACSF descriptor spaces, {as well as further local atomic density descriptors, such as the atomic cluster expansion (ACE) representation, have been shown to be compressible without significant loss of information, improving computational efficiency} \cite{Darby2022, Zeni2021}.

The resulting $DII$ for various numbers of ACSFs can be seen in Fig.~\ref{fig:force_prediction}A.
With both greedy and L$_1$ regularized selection, we find that the optimized $DII$ asymptotically approaches an optimal value with growing number of non-zero features.
However, even relatively small feature spaces with $\sim$ 10-30 non-zero features have low $DII$ values, making effective feature selection possible. We validate the selected features and their weights on validation sets of atomic environments of equal size as the training set. The resulting $DII$s are slightly higher but mostly comparable to the training $DII$s, showcasing the robustness and transferability of the results.
As a sanity check for our selection, we also show that randomly selected feature sets have a significantly higher $DII$ than optimized sets, meaning they are less informative about the ground truth space (Fig.~\ref{fig:force_prediction}A gray).

To show that the features selected by $DII$ are indeed physically relevant, we report in Fig.~\ref{fig:force_prediction}B the root-mean-square error (RMSE) of atomic forces for Behler-Parrinello MLPs using ACSF subsets of different sizes (n$_{\text{ACSF}} \in \{10, 18, 25, 38, 50, 176\}$).
We find that MLPs with features selected by L$_1$ regularized $DII$ optimization outperform random input features for all tested numbers of input features n$_{\text{ACSF}}$.
The difference in prediction accuracy is most pronounced at small n$_{\text{ACSF}}$, where it is least likely that random selection chooses meaningful features.
After n$_{\text{ACSF}} \approx 20$ input features, the optimized subsets reach an accuracy of < 100 meV, which is on par with the original MLP trained on these data \cite{Cheng2018}. {Compressions of local atomic density representations for machine learning potentials have also previously been shown to require a minimum set size of 10-20 PCA features, since further compression fails to faithfully preserve the geometric relationships between data points and leads to increased prediction errors} \cite{Zeni2022}.
With n$_{\text{ACSF}}=50$ input features, the MLP performs roughly equally well to using the full data set, while having less than half the run-time (Fig.\ref{fig:force_prediction}C).
This shows that $DII$ can be used to select features for downstream tasks such as energy and force fitting in MLPs, by optimizing for a complex ground truth and finding a space with fewer but optimally weighted features that contain the same information.

\section*{Discussion}
This work presents the Differentiable Information Imbalance, $DII$, designed to automatically learn the optimal distance metric $d^A$ over a set of input features. The metric reproduces the neighborhoods of the data points as faithfully as possible according to a ground truth distance $d^B$.
Here $d^A$ is defined as the Euclidean distance, and the optimization parameters are weights that scale individual features, such that the presented $DII$ is an automatic {and universal} feature selection and weighting algorithm. 

{While many other methods are restricted to single variable outputs as “labels” or “targets”, $DII$ can handle any dimensionality of input and output. Continuous and discrete data is supported and the method can be used in a supervised and unsupervised manner. The weights are optimized automatically, and by using the values of the $DII$ as a quality measure one can  compare the information content of several  feature sets, and select the sets corresponding to the lowest $DII$ for each number of features, such as in Suppl. Fig. S6.} It is one of very few filter methods that account for feature dependencies but do not rely on explicit feature subset evaluation \cite{Urbanowicz2018}. 

In illustrative examples where the optimal feature weights are known, we showed that the $DII$ can reliably find the correctly weighted ground truth features out of high-dimensional input spaces. The behavior of the $DII$ as a function of the subset size appears to be anti-correlated with the cosine similarity between ground truth and optimized weights. This implies that the $DII$ value can be used for assessing the quality of the selected feature subsets when the actual ground truth weights are unknown.
The weighted feature sets as provided by $DII$ optimization have a higher cosine similarity to the ground truth than sets derived from two other feature selection classes, relief-based algorithms (RBAs) {\cite{Urbanowicz2018bioinformatics}} and decision tree regressions.

We further applied the method to analyze a molecular dynamics (MD) simulation of a biomolecular system. Extracting a small subset of informative collective variables (CVs) from a pool of many candidate CVs from a MD trajectory is a general problem with both practical and conceptual benefits, including using such CVs in enhanced sampling techniques and obtaining an interpretable  description of the free energy landscape.
For the peptide CLN025, the selected CVs are the first two principle components ($3.5$ PC1, $4.7$ PC2) and the radius of gyration ($1.0$ RGYR).
Applying clustering in the space of these three scaled CVs leads to the correct identification of the $\beta$-pin state and collapsed denatured state of CLN025, in accordance with the clusters built from a much larger feature space, which includes all heavy atom distances. The reduced space clusters are highly meaningful with a $89\%$ overall cluster purity towards the extended space clusters, while reduced variable spaces built from clustering results of the decision tree regression lead to lower cluster purities. Tests of uncorrelated parts of the MD trajectory show great consistency of the results, highlighting the robustness of the method.

In a second application, our method successfully selects highly informative subsets of input features for training a Behler-Parrinello machine learning potential that achieves optimal performance in terms of the mean absolute error of force and energy.
We find that using just 50 informative ACSF descriptors selected by our approach, instead of 176, significantly reduces the MLP’s computational cost, cutting the runtime by one third while maintaining nearly the same accuracy.

The $DII$ is not necessarily a simple monotonic function of the number of non-zero features post-optimization (cardinality). In some cases, the selection of additional features can introduce noise or redundancies that can negatively impact the description of the ground-truth space.
Furthermore, if the optimal non-zero features for a two-dimensional description are, say, X3 and X61, the optimal features for a three dimensional description could be completely different, say X5, X9 and X44. The $DII$ is hence also not necessarily a submodular function of the number of features.

To extract small subsets of features from high dimensional input data, we implemented two different sparsity inducing heuristics:  L$_1$ regularization and greedy backward optimization. Greedy algorithms have previously been shown to be a fast and effective alternative to convex L$_1$ regularization in sparse coding \cite{Huamin2015}, and work even if the problem is only approximately submodular \cite{halabi2022}. When a feature space is very large, greedy backward optimization will lead to long calculations and L$_1$ regularization becomes more suitable. Both heuristics are able to find relevant results in the examples presented here.

Like RBAs {\cite{Urbanowicz2018}}, also $DII$ has a computational complexity of $\mathcal{O}(N^2 \cdot D)$, where $N$ is the number of points and $D$ is the number of features. However, by applying a simple subsampling trick (see \nameref{sec:rowtrick}), the computational complexity reduces up to $\mathcal{O}(N \cdot D)$ with a degradation of the accuracy which is barely detectable (Suppl. Fig. S3).

The requirement of a ground truth reference space could pose a difficulty to some applications. 
In MD simulations, all heavy atom distances are a good, translationally invariant alternative to the set of all atomic positions, if one wants to completely encode the conformation of a molecule. In other cases, if no independent ground truth is known or a lower-dimensional subspace is desired, the full space could be used as ground truth. This approach could be employed, for example, for large gene sequencing data with thousands of features and just hundreds of data points. In this fashion, the method acts as an unsupervised feature selection filter. An open question in this case is the relative weighting of the ground truth features.

Furthermore, even though the method can be applied to any data set, it is most suitable for continuous features. A limitation is given by ground truth metrics with many nominal or binary features, which can lead to a degenerate ground truth rank matrix, making the optimization more difficult.

The Differentiable Information Imbalance introduced in this work could have relevant implications in a wide range of distance-based methods, such as k-NN classification, clustering, and information retrieval.
The approach could also be used to identify how much information original features carry compared to otherwise not-interpretable transformations such as UMAP \cite{McInnes2018} or highly non-linear neural network representations, by optimizing the original features towards such representations. 
Defining a new feature by combining several input features through a (possibly nonlinear) function might bring to even more compressed and informative representations, although this could reduce interpretability.
$DII$ has also potential applications beyond feature selection with automatic weighting.
Specifically, constructing a distance space $d^A(\boldsymbol{w})$ with a more expressive functional form, compared to the one used in this work, opens up to applications in fields such as dimensionality reduction \cite{maaten2009dimensionality,glielmo2021unsupervised} and metric learning \cite{bellet2015metric}.

The Differentiable Information Imbalance has been implemented in the Python library DADApy \cite{dadapy2022} and is well-documented \cite{Readthedocs}, including a tutorial for ease of use. This accessibility allows for a wide audience to explore further use cases and limitations effectively.

\section*{Methods}

\subsection*{Adaptive softmax scaling factor $\lambda$}
\label{sec:adalambda}
Qualitatively, the scaling factor $\lambda$ in the softmax coefficient $c_{ij}(\lambda, \boldsymbol{w})$ defines the size of the neighborhoods in the input space $d^A(\boldsymbol{w})$ used for the rank estimation. 
Since $\lambda$ is the same for every data point, regardless of whether the point is an outlier or within a dense cloud, this factor mainly decides how many neighbors are included in dense regions of the data manifold.
Importantly, choosing $\lambda$ too small makes the optimization less efficient, as in the limit $\lambda\rightarrow 0$ the derivative of the $DII$ (see Eq. (\ref{eq:derivative})) can be shown to vanish for almost all values of the parameters $\boldsymbol{w}$.

To automatically set $\lambda$, we take the average of two distance variables, $\hat{d}_{\,\text{min}}^{A}$ and $\hat{d}_{\,\text{avg}}^{A}$, which heuristically define the “small distance” scale in space $d^A$.
Both of these numbers are based on $\hat{d}_{i}^{A}$, here denoting the difference between 2nd and 1st nearest neighbor distances for each data point $i$, $\hat{d}_{i}^{A}= d_{ik}^{A} - d_{ij}^{A}$, where $r_{ij}^{A} = 1$ and $r_{ik}^{A} = 2$:
\begin{subequations}
\begin{align}
	\hat{d}_{\,\text{min}}^{A} &:= \min_i \hat{d}_{i}^{A}\,, \\
	\hat{d}_{\,\text{avg}}^{A} &:= \frac{1}{N}\sum_i \hat{d}_{i}^{A}\,.
\end{align}
\end{subequations}
Setting $\lambda$ to the average of $\hat{d}_{\,\text{min}}^{A}$ and $\hat{d}_{\,\text{avg}}^{A}$ at each step of the $DII$ optimization has proven to enhance both the speed and stability of convergence.
Indeed, using differences between nearest neighbor points to determine $\lambda$ is more robust than using nearest neighbor distances directly, as in high dimensions first-, second- and higher-order neighbor distances tend to be very similar on a relative scale \cite{beyer1999whenisnn,hinneburg2000whatisnn}.

\subsection*{Invariance property of the $DII$}
\label{sec:invar}
In the limit $\lambda\rightarrow 0$, the $DII$ defined in Eq. (\ref{eq:diff_II}) is invariant under any global scaling of the distances in space $A$, $d_{ij}^A\mapsto |c|\,d_{ij}^A$ with $c\in\mathbb{R}$. Similarly, in the small $\lambda$ regime, $DII(d^A(\boldsymbol{w})\rightarrow d^B)$ is invariant under any uniform scaling of the weight vector, $\boldsymbol{w}\mapsto c\,\boldsymbol{w}$, if $d^A(\boldsymbol{w})$ is built as the usual Euclidean distance in the scaled feature space.
This property can be easily verified by observing that the softmax coefficients $c_{ij}$ can be replaced by $\delta_{1,r_{ij}^A}$ when $\lambda\rightarrow 0$, and the ranks $r_{ij}^A$ are invariant under a global scaling of the distances $d_{ij}^A$.
The same invariance holds even for $\lambda > 0$ if $\lambda$ is chosen adaptively (see \nameref{sec:adalambda}), as in the adaptive scheme a global scaling of the distances $d_{ij}^A$ implies a scaling of $\lambda$ by the same factor, which leaves the $c_{ij}$ coefficients untouched.

\subsection*{Optimization of the $DII$}
\label{sec:opti}
The optimization of the $DII$ is implemented in
\texttt{FeatureWeighting.\-return\_\-weights\_\-optimize\_dii} in DADApy by gradient descent utilizing the analytic derivative of the $DII$. The default value of the initial feature weights is the inverse standard deviation of each feature. 
{Pseudocodes of the $DII$ optimization algorithms are provided in the Suppl. Inf. (section ``Pseudocodes'').}

\subsubsection*{Learning rate decay}
We employ two different schemes of learning rate decay, (1) cosine learning rate decay and (2) exponential learning rate decay. When both schemes are evaluated, we select the solution with lower $DII$ among those found with the two schemes. In the first scheme, the learning rate is updated according to $\eta^k = 0.5 \eta^0 \cdot (1 + \cos(\frac{\pi k}{n_{\text{epochs}}}))$, where $k$ denotes the training epoch, $\eta^0$ the initial learning rate, and $n_{\text{epochs}}$ the total number of epochs in the training. The exponential decay follows $\eta^k = \eta^0 \cdot 2^{\frac{-k}{10}}$. This schedule cuts the learning rate by half every 10 epochs. While the cosine decay leads to optimal results in the absence of L$_1$ regularization, or for weak regularization, the exponentially decaying learning rate is especially suited for high L$_1$ regularization. In both schemes, ``GD clipping'' is used, as described hereafter in the section on L$_1$ regularization.

\subsubsection*{L$_1$ regularization}
\label{sec:l1}
This method is implemented in DADApy in \texttt{FeatureWeighting.\-return\_\-weights\_\-optimize\_\-dii} when a L$_1$ penalty different from $0$ is chosen, and several different L$_1$ values are screened in \texttt{FeatureWeighting.\-return\_\-lasso\_\-optimization\_\-dii\_\-search}.
Optimizing the $DII$ with respect to the feature weights while simultaneously introducing sparsity, i.e. limiting the number of features used, can be considered a convex optimization problem of the form:
\begin{equation}
    \min_{\boldsymbol{w} \in \mathbb{R}^D} \big( f(\boldsymbol{w}) + p\, \Omega(\boldsymbol{w}) \big),
\label{minopt}
\end{equation}
where $f:\mathbb{R}^D \rightarrow \mathbb{R}$ is a differentiable function such as $DII \left(d^A(\boldsymbol{w})\rightarrow d^B\right)$, at least locally convex, and $\Omega:\mathbb{R}^D \rightarrow \mathbb{R}$ is a sparsity-inducing, non-smooth, and non-Euclidean norm with penalization strength $p$ \cite{Bach2011}.
We use the L$_1$ norm, $\Omega(\boldsymbol{w}) = \sum^D_{\alpha=1} |w^{\alpha}|$ (also called lasso regularization): 

\begin{equation}\label{eq:diff_II_l1}
    \min_{\boldsymbol{w} \in \mathbb{R}^D} \left( DII + p\, \Omega(\boldsymbol{w}) \right) = \\ 
    \min_{\boldsymbol{w} \in \mathbb{R}^D} \left( \frac{2}{N^2}\,\sum_{\substack{i,j=1 \\ (j\neq i)}}^N c_{ij}(\lambda,d^A(\boldsymbol{w}))\,r_{ij}^B 
    + p\,\sum^D_{\alpha=1} |w^{\alpha}|
    \right) 
\end{equation}

The L$_1$ norm has the shortcoming that in $N \ll D$ setting, with very few samples but many dimensions, a maximum of $N$ variables can be selected.
The L$_1$ regularization tends to select just one variable from a group of correlated variables and ignore the others \cite{Zou2005}, which helps building optimal groups of maximally uncorrelated features (see Supplementary Information in ref. \cite{Wild2024}).

Naive gradient descent with L$_1$ regularization usually does not produce sparse solutions, as a weight becomes zero only when it falls directly onto zero during the optimization \cite{Tsuruoka2009}. This is very unlikely with most learning rate regimes. Instead, we employ the two-step weight updating approach also known as ``GD clipping'' \cite{Tsuruoka2009}:

\begin{equation}
    \begin{aligned}
     w_{t+\frac{1}{2}}^{\alpha} =
     & \,w_{t}^{\alpha} -  \frac{\partial DII\left(d^A(\boldsymbol{w})\rightarrow d^B\right)}{\partial\, w^\alpha} \\
     & \textrm{\textbf{if }} w_{t+\frac{1}{2}}^{\alpha} > 0 \textrm{\textbf{ then }} w_{t+1}^{\alpha} =  \max(0, w_{t+\frac{1}{2}}^{\alpha} - \eta\, p)  \\
     & \textrm{\textbf{if }} w_{t+\frac{1}{2}}^{\alpha} < 0 \textrm{\textbf{ then }} w_{t+1}^{\alpha} =  |\min(0, w_{t+\frac{1}{2}}^{\alpha} + \eta\, p)|
    \end{aligned}
    \label{eq:tsuruoka}
\end{equation}

Here, $p$ denotes the L$_1$ penalty strength, and $t$ is the epoch index. First, the update is performed only with the GD term, which may result in a change of sign for the weight. Subsequently, the L$_1$ term is applied, shrinking the weight magnitude. If this shrinkage would change the weight's sign, the weight is instead set to zero. Since the $DII$ is sign invariant, all weights are kept positive during the optimization.

\subsubsection*{Backward greedy optimization}
This approach is implemented in DADApy in \texttt{FeatureWeighting.\-return\_\-backward\_\-greedy\_\-dii\_\-elimi\-nation}. 
It starts with a standard optimization run using all the $D_A$ features of the input space. From the solution of the first optimization, the feature corresponding to the smallest weight is discarded (set to zero), and a new optimization with $D_A - 1$ features is carried out.
This procedure is iterated until the single most informative feature is left.
The greedy backward approach is an alternative to the L$_1$ regularization and is applicable to moderately large data sets with $D_A \lesssim 100$ features and $N \lesssim 500$ data points, since the computational complexity scales linearly with the number of features.

\subsection*{A linear scaling estimator of the $DII$}
\label{sec:rowtrick}
The $DII$ scales quadratically with the number of points $N$, with a computational complexity of $\mathcal{O}(N^2 \cdot D)$, where $D$ is the number of features.

The computational time can be dramatically decreased by subsampling the rows of the matrices $r_{ij}$, $d_{ij}$ and $c_{ij}$ appearing in Eq. ({\ref{eq:diff_II}}), reducing them to a rectangular shape $N_{\text{rows}}\times N$ (with $N_{\text{rows}} < N$) (see Suppl. Inf. "Tests of scalability and robustness"). This subsampling is performed only once at the beginning of the training, so that the rectangular shape of such matrices is kept fixed during all the $DII$ optimization.
If the $DII$ is written as the average of $N$ conditional ranks,
\begin{equation}
    DII\left(d^A(\boldsymbol{w})\rightarrow d^B\right) = \frac{2}{N}\frac{1}{N}\,\sum_{i=1}^N \bigg( \sum_{\substack{j=1 \\ (j\neq i)}}^N c_{ij}(\lambda,d^A(\boldsymbol{w}))\,r_{ij}^B\bigg)  = \frac{2}{N} \langle r^B | r^A \approx 1 \rangle\,,
\end{equation}
the subsampling is equivalent to replacing $1/N\sum_{i=1}^N$ with $1/N_{\text{rows}}\sum_{i=1}^{N_{\text{rows}}}$. This means computing the average of $N_{\text{rows}}$ conditional ranks instead of $N$.
Different schemes to set $N_{\text{rows}}$ result in different scaling laws of the algorithm with respect to $N$. Setting $N_{\text{rows}}$ to a fraction of $N$ (green curve in Suppl. Fig. S3A, $N_{\text{rows}} = N/2$) brings to a quadratic scaling with a smaller prefactor, while sampling a fixed number of points $N_{\text{rows}}$ independently of $N$ (red curve, $N_{\text{rows}} = 100$) brings to a linear scaling $\mathcal{O}(N \cdot D)$. In the latter case we observe a striking reduction of the runtime, while the accuracy of the recovered weights is almost perfectly preserved (Suppl. Fig. S3B).

\subsection*{Extraction of collective variables from the CLN025 MD simulation}
\label{sec:cvs}
All collective variables were extracted from the MD simulation using PLUMED 2 \cite{PLUMED2014}.
The ground truth pairwise heavy atom distances were computed using the ``DISTANCE'' CV on all pairs of non-hydrogen atoms. 
The radius of gyration was obtained with the ``GYRATION'' CV and the C$_{\alpha}$ atoms. 
The number of hydrophobic contacts were calculated using the ``COORDINATION'' CV (R$_0$=0.45) and using the amino acids THR, TRP, and TYR of CLN025, and sidechain carbons not directly bonded with an electronegative atom.
The number of hydrogen bonds was also calculated using the ``COORDINATION'' CV (R$_0$=0.25). For backbone H-bonds and sidechain H-bonds only hydrogens and oxygens of the backbone and the sidechain were considered, respectively, while for the sidechain-to-backbone interactions, the cross of these were considered.
For the quantification of the alpha-helical content and the anti-parallel beta sheet content, the CVs ``ALPHARMSD'' and ``ANTIBETARMSD'' were used with all residues of the peptide.
For the principle components PC1, PC2 and the PCA residual, first a pdb file containing the average structure of the trajectory and the two first principle directions was created using the CVs ``COLLECT\_FRAMES\_ATOMS'' with all heavy atoms, and ``PCA'' using the previous output and optimal alignment. Subsequently, each frame of the trajectory was projected onto the two principle components referenced in the pdb file using ``PCAVARS''. 

\subsection*{Block cross validation of CLN025}
\label{sec:crossval}
To account for the equilibration of the system, the first $\sim 15$ ns of the trajectory were discarded throughout the analysis (1,580 of 41,580 trajectory frames).
Block cross validation (Fig.~\ref{fig:CLN1}A) was carried out by splitting the remaining frames into 4 consecutive blocks. The training blocks were built by subsampling each block to every 7th frame to de-correlate, leaving 1428 points per training block. 
The optimal tuple and weight results from each training block were used to calculate the $DII$ in 21 validation sets built from the remaining three blocks (repeatedly subsampling each block with stride 7, starting from frames 1 to 7), totaling 84 validation sets. 

\subsection*{ACSF and SOAP descriptors}
\label{sec:mlp}
The systems for creating ACSF and SOAP descriptors are based on 1593 liquid H$_2$O structures whose forces and energies were found using DFT via the CP2K \cite{Lippert1999} package with the revPBE0-D3 functional.
We use the DScribe Python package \cite{dscribe, dscribe2} to calculate SOAP and ACSF descriptors from the atomic positions. The data points were chosen as follows: The 1593 structures (with 64 H$_2$O molecules each) yielded 192,000 atomic environments, from which a subset of $\sim$350 was sampled to reduce the computational time of feature selection.
The ACSF descriptors were constructed on a grid of hyperparameters (G2: $\eta \in [10^{-3}, 10^{0.5}]$ logspace n$_\eta=15$, $R_S=0$, G4: $\eta \in [10^{-3}, 10^{0.5}]$ logspace n$_\eta=6$, $\zeta \in \{1, 4\}$, $\lambda \in [-1, 1]$ linspace n$_\lambda=4$, $R_S=0$), resulting in 176 (+2 cutoff functions) different features for each atomic environment.
The 546 SOAP descriptors were selected with n$_{\text{max}}=6$, l$_{\text{max}}=6$ and a cutoff radius of 6\AA.

The optimization of ACSF with respect to the ground truth of SOAP is carried out starting from $\gamma_i = 1 \hspace{1mm} \forall \hspace{1mm} i \in [1, 176]$.

\section*{JAX version of DII}
In order to benefit from modern machine learning GPU-based calculation speed, a GPU-compatible implementation written with the JAX library {\cite{jax2018github}} is also provided within the same package.

\section*{Data availability}
The data generated by feature selection in this study have been deposited on OSF at the following URL: \\ \hyperlink{https://osf.io/swtg5}{https://osf.io/swtg5}. Source data are provided with this paper. The processed molecular dynamics and H$_2$O structure data are also available at OSF. The data files necessary for carrying out all analyses and source data are available at the same OSF URL.

\section*{Code Availability}
The Python code to replicate and extend our study is available on GitHub at the following URL: \\ \hyperlink{https://github.com/sissa-data-science/DADApy}{https://github.com/sissa-data-science/DADApy} under the Apache License 2.0. The according documentation can be found in ref. \cite{Readthedocs}. The code at the time of publishing can be built under the Apache License 2.0 from: \hyperlink{https://doi.org/10.5281/zenodo.14277899}{https://doi.org/10.5281/zenodo.14277899} \cite{Wodaczek2024}.

\printbibliography

\section*{Acknowledgements}
The authors thank Dr. Matteo Carli for providing the CLN025 replica exchange MD trajectory and Matteo Allione for the fruitful discussions connected with the idea of the linear scaling estimator. This work was partially funded by NextGenerationEU through the Italian National Centre for HPC, Big Data, and Quantum Computing (Grant No. CN00000013 received by A.L.).
A.L. also acknowledges financial support by the region Friuli Venezia Giulia (project F53C22001770002 received by A.L.).

\section*{Author contributions}
Concept: A.L., R.W., V.D.T., F.W. and B.C. Model and algorithm development: R.W., V.D.T., F.W., B.C. and A.L. Code implementation: R.W., F.W. and V.D.T. Figures: R.W., F.W. and V.D.T. Writing: R.W., V.D.T., F.W., B.C. and A.L.

\section*{Competing interests}
The authors declare no competing interests.

\end{document}


\maketitle

\section{Benchmark with other feature selection algorithms}

\subsection{Gaussian random variables and their products}
We benchmark the $DII$ approach against other feature selection algorithms on the high-dimensional benchmark example with known ground truth described in the main text. The input is given by 285 features which are monomials up to order three of 10 Gaussian random variables. The ground truth are ten of these features chosen at random and scaled with various weights (Table 2 of the main text). The task for the feature selectors is, given the 10-dimensional ground truth, finding these ten features out of the 285. We only compare with feature selection methods that can assign weights to features. 

\subsubsection*{Relief-based algorithms (RBAs)}
We first consider the relief-based algorithms (RBAs) RReliefF and MultiSURF, implemented in scikit-rebate \cite{Urbanowicz2018bioinformatics}, which support a continuous ground truth \cite{Urbanowicz2018}. The RBAs stand out among filter methods because they can assign feature weights. They output weights between -1 (most irrelevant) and 1 (most relevant) for each feature, but importantly only work with one-dimensional ground truth. 
We ran the algorithms vs. each feature of the ground truth separately and (\textit{a}) summed all resulting weights that scale the input features or (\textit{b}) set all resulting weights except the largest to zero and summed these sparse vectors. Then we calculated the cosine similarity (overlap) with the 285-dimensional ground truth vector (all weights zero except the ten relevant weights, which are set to their value).

\begin{figure}[!ht]
    \centering
    \includegraphics[width=.9\textwidth]{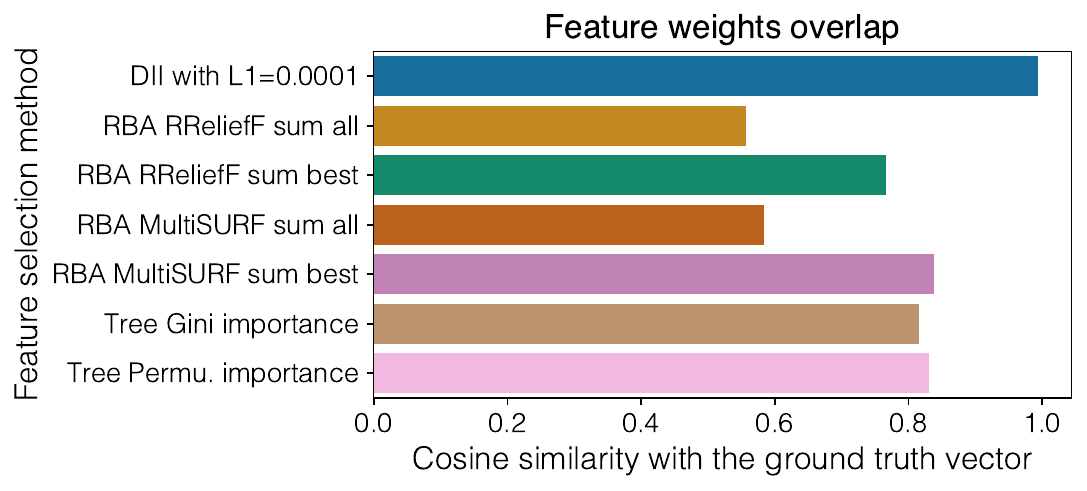}
    \caption{\textbf{Overlap of the weight vectors resulting from different feature selection and weighting methods with the ground truth weight vector, calculated as the cosine similarity.} The Differentiable Information Imbalance ($DII$) result is shown for a L$_1$ regularization of $0.0001$. For the Relief-based algorithms (RBA) RReliefF and RBA MultiSURF, ``sum all'' refers to the sum over the ten individual optimizations (for each ground truth feature as univariate label) of the resulting weight vectors, while ``sum best'' means the same sum over individual optimizations, but setting all weights to zero  except the largest one in each feature vector. 
    The decision tree ('Tree') results refer to the Gini and Permutation (Permu.) importance selected feature weights as provided by the decision tree regression aimed at the full ten-dimensional target space.
    Source data are provided as a Source Data file.}
    \label{fig:overlap_m}
\end{figure}

Both variants (RReliefF and MultiSURF) correctly assign the dominant weight to the respective ground truth feature for nine out of the ten ground truth features. Only a feature with a comparatively smaller variance was not reliably detected. Therefore, the algorithm performs well in filtering relevant features. However, since each gt feature has to be evaluated separately, there is no reliable relative weighting of the 10 dominant features. The final weight vector used to calculate the cosine overlap was therefore constructed according to \textit{a} (orange and bown-red in Fig. \ref{fig:overlap_m}), or according to \textit{b} (green and light purple in Fig. \ref{fig:overlap_m}). The resulting cosine similarities are $0.58$ and $0.84$ for MultiSURF, and $0.56$ and $0.77$ for RReliefF, for \textit{a} and \textit{b} respectively. The importance cutoff, \textit{i.e.} setting all but the largest weight to zero, consistently led to better results for the RBAs.

\subsubsection*{Decision tree regression}
The second benchmark was carried out using the decision tree regressor (\texttt{sklearn.tree.DecisionTreeRegressor}) implemented in the scikit-learn package \cite{scikit-learn}, an embedded method which can handle multi-target contiuous ground truths.

Tuning the algorithm for various error criterions and splitters, the defaults (\texttt{criterion='squared\_error'}, \texttt{splitter='best'}) performed best. The feature importances were derived in two ways: 1) Gini importance inherent to the DecisionTreeRegressor, and 2) Permutation importance for for feature evaluation \cite{Breiman2001} (\texttt{sklearn.inspection.permutation\_importance} with \texttt{n\_repeats} tuned to $10$). The latter is expected to be more robust for high cardinality features with many unique values, like in the test example here. We use the Gini importance and the Permutation importance weight vectors to calculate the cosine similarities of 0.82 and 0.83 with the ground truth, respectively. The six largest ground truth features are among the top ten weighted features according to both, Gini and Permutation, even if not in the correct order. 

\subsection{Identifying optimal collective variables for the peptide CLN025}
\label{sec:CLNsupp1}
Because of the good performance of decision tree regression on the previous example and its ability to handle high-dimensional ground truth spaces, we apply this feature selection algorithm also to the molecular dynamics (MD) trajectory of the peptide CLN025. In this case the ground truth corresponds to the 4,378  heavy atom distances which are used to define the target distance. The input variables on which we perform the feature selection are the collective variables (radius of gyration (RGYR), anti-$\beta$-sheet content, $\alpha$-helical content, number of hydrophobic contacts, principal component 1 (PC1), principal component 2 (PC2), principal component residuals, the number of hydrogen bonds in the backbone, in the side chains, and between the backbone and side chains; for details on the CVs see Methods).

The best three variables found using the Gini importance weights are: $0.29$ anti-$\beta$-sheet content, $0.25$ PC1, $0.1$ PC2; using the permutation importance they are: $1.27$ PC1, $1.04$ anti-$\beta$-sheet content, $0.97$ PC2.

Notice that in this test the ground truth variables and the input variables are different sets. Therefore, it is not possible to assess directly the quality of the result by computing a cosine similarity. We instead evaluate the quality of the features, which were selected by the decision tree regression, following the same procedure as used in the main text, by cluster purities and other clustering numbers.
Decision tree regression brings to overall cluster purities compared to the full space clusters of 0.55 for Gini importance and 0.63 for the permutation importance. The cluster center for the collapsed loop is in both cases the same, and the collapsed loop cluster is bigger than the native $\beta$-pin cluster, which contradicts the results obtained by full space clustering, where the $\beta$-pin cluster is the largest (with most frames) and the lowest free energy cluster. Visual comparison confirms that the cluster centers of the two metastable states, the $\beta$-hairpin and the collapsed loop, derived from the decision tree regression feature selection, are less similar to the full space cluster centers than the DII-derived CV space in Fig. 2 of the main text. In particular, even if the backbone is in a similar conformation, the side chains are arranged in a different manner. 

\begin{figure}[!ht]
    \centering
    \includegraphics[width=.99\textwidth]{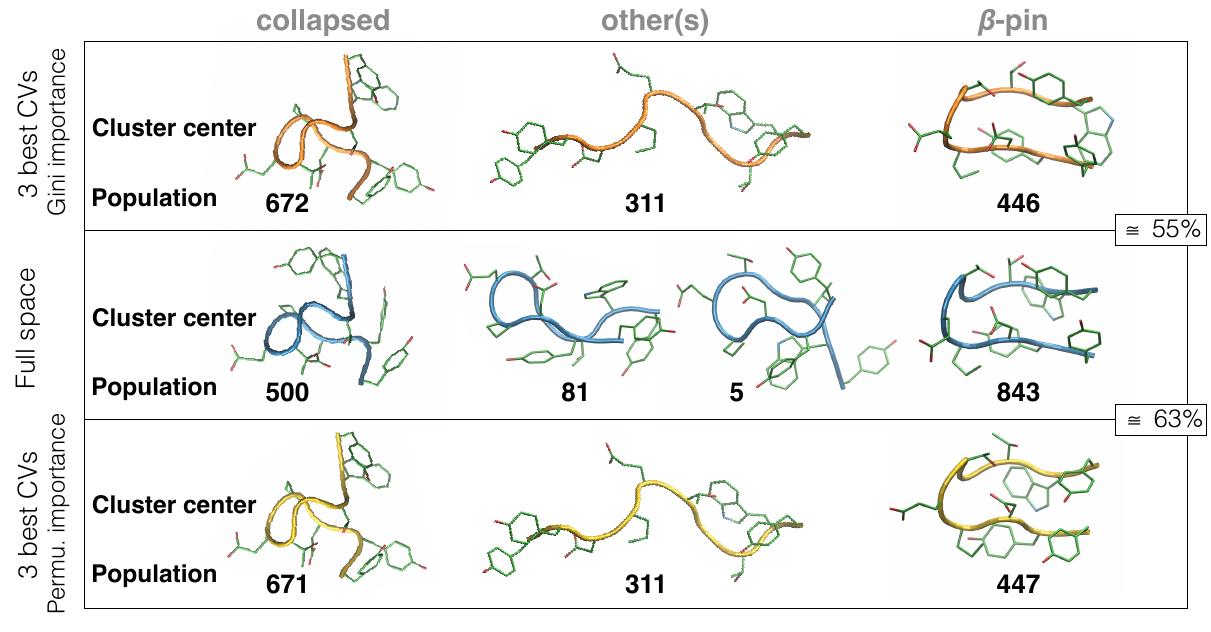}
    \caption{\textbf{Cluster centers and populations of CLN025, derived from three different sets of collective variables.} Density Peak Clustering was used. Blue is the clustering done in the full space of 4,378 variables, orange in the space of the three best variables as selected and weighted by the Gini importance of the Decision Tree Regression model, and yellow as selected and weighted by the Permutation importance. The numbers on the right are the cluster purities with respect to the full space, measured as the fraction of points (trajectory frames) that belong to the same cluster in both representations.}
    \label{fig:tree_clusters}
\end{figure}

\newpage
\section{Tests of scalability and robustness}
In the following we test how the $DII$ performs for various number of data points in terms of quality of the results and in terms of computational complexity, or runtime.

\subsection{Products of Gaussian random variables}

\begin{figure}[!ht]
    \centering
    \includegraphics[width=.99\textwidth]{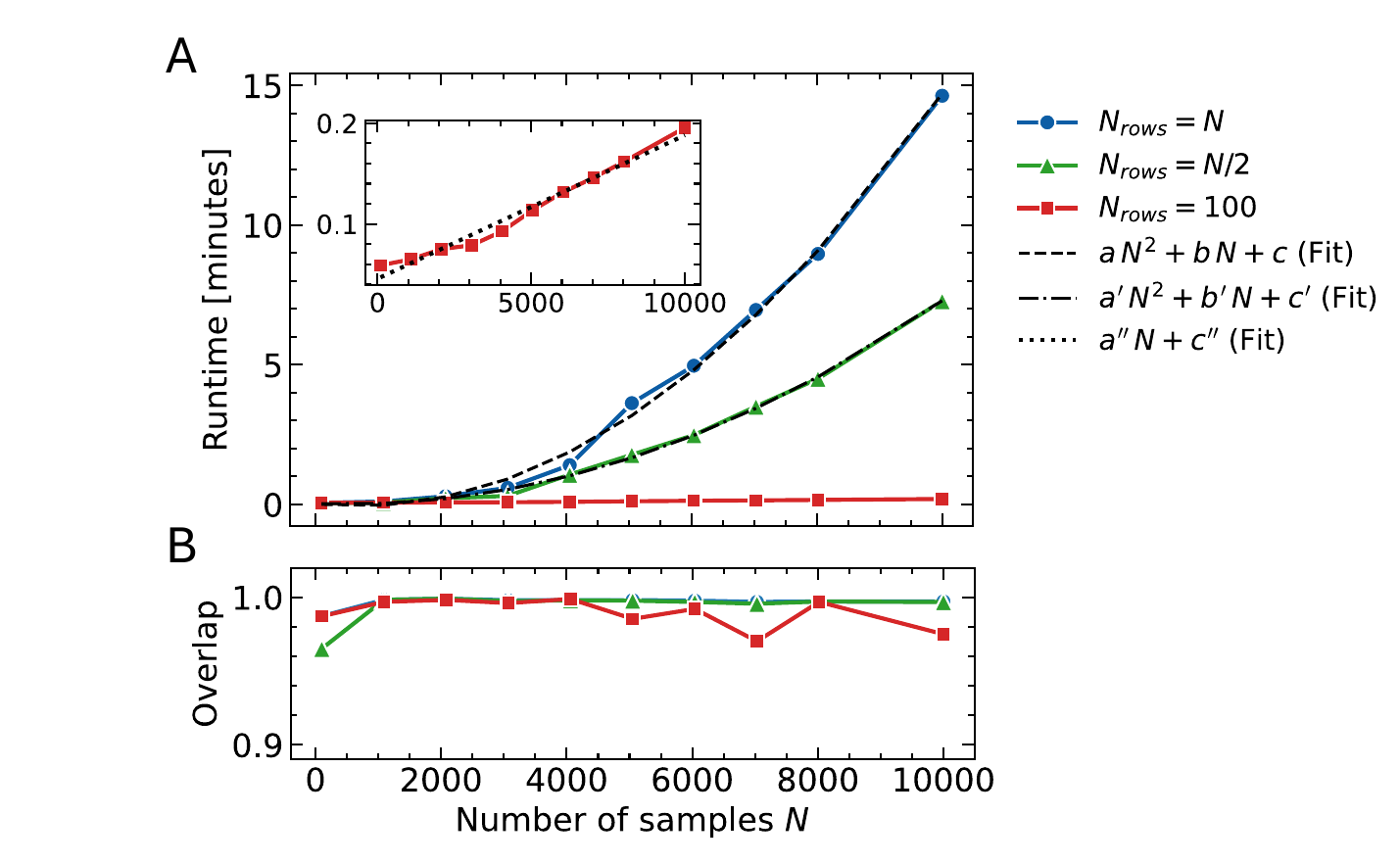}
    \caption{\textbf{Scalability of the algorithm tested on monomials of Gaussian random variables. A}: Runtime of the $DII$ optimization as a function of the data set size $N$, using the 285 monomial features in the input space and 5 non-zero features in the ground-truth space. Each optimization is carried out with 1000 epochs, setting the L$_1$ penalty to $10^{-3}$.
    Different colors show how the runtime is affected by different row subsampling schemes: No subsampling (blue), linear subsampling in the number of points (green), constant $N_{\text{rows}}$ (red).
    The dashed and dotted black lines show least square fits with the expected scaling laws.
    \textbf{B}: Overlap between learned weights and ground truth weights, computed as the cosine similarity between the two weight vectors.
    Source data are provided as a Source Data file.}
    \label{fig:scalability}
\end{figure}

We test the scalability of our algorithm with respect to the number of points $N$ used to perform the minimization of the $DII$ on the example of 285 monomials (Fig. 1B of the main text). We construct the ground-truth distance $d_B$ by multiplying 5 features with non-zero weights.
In Fig. \ref{fig:scalability}A we show the runtime for a single optimization of the $DII$ as a function of $N$ (ranging from 100 to 10000).
As a quality validation measure, we report in Fig. \ref{fig:scalability}B the overlap (cosine similarity) between the learned and the ground-truth weights. 
The tests have been performed using the JAX implementation of the algorithm on a GPU nVidia TU104GL [Quadro RTX 5000].

The standard algorithm presented in the main text ($N_{\text{rows}}=N$, blue line) scales quadratically with the number of points $N$.
Indeed, the main steps of the algorithm (computing the $DII$ and its gradient, Eqs. (2) and (6) in the main text) can be performed in $\mathcal{O}(N^2 \cdot D)$ operations, as they both involve a double sum over the rows (index $i$) and the columns (index $j$) of the $N\times N$ matrices in the equations. $D$ is the number of features.
We notice that the additional sums over index $m$ (denominator of the $c_{ij}$ coefficients, Eq. (3), and second term in the gradient, Eq. (6)) do not depend on index $j$ and can therefore be precomputed for each index $i$, avoiding three nested loops.

The computational time can be dramatically decreased by performing the sum over $i$ only on a fixed subset of points, as explained in the Methods section of the main text. This leads to a striking reduction of the runtime, and in the best case to a linear scaling $\mathcal{O}(N \cdot D)$ in the number of data points. Notably, the accuracy of the recovered weights is almost preserved (Fig. \ref{fig:scalability}B).

\subsection{The peptide CLN025}
\begin{figure}[!ht]
    \centering
    \includegraphics[width=.65\textwidth]{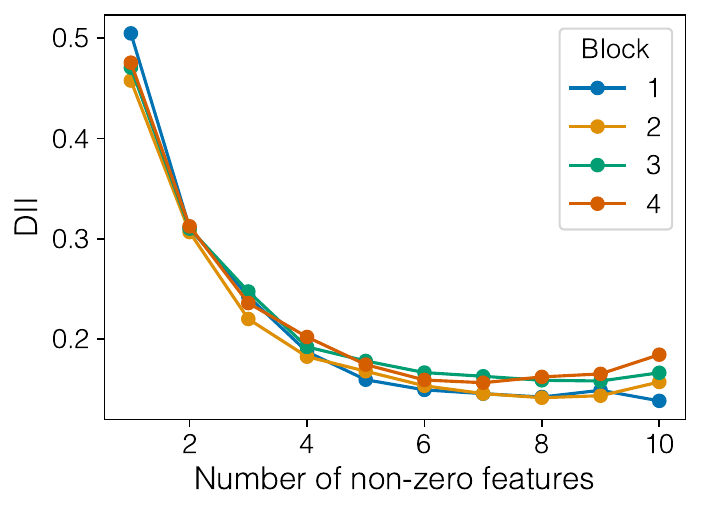}
    \caption{\textbf{Block analysis of CLN025 to test robustness.} The Differentiable Information Imbalance ($DII$) for each of four uncorrelated trajectory blocks of CLN025 with 1429 frames each, \textit{vs.} the number of selected (non-zero) features of a total of ten input features.
    Source data are provided as a Source Data file.}
    \label{fig:DII4blocks}
\end{figure}

The second robustness test is carried out by cutting the MD trajectory of the peptide CLN025 into 4 blocks of 10000 frames each and subsampling every 7th frame, to create 4 uncorrelated data sets of 1429 points each - the same size of data set as in the analyses in the main paper. For each of these sets, we found the optimal number of non-zero features and optimal relative weights of these features with $DII$-optimization, using the full space consisting of 4,378  heavy atom distances as target. Fig. \ref{fig:DII4blocks} shows that the DII decreases steadily until approximately 5 to 6 features, when the information content reaches its maximum. More than seven or eight features increase the $DII$, since noise seems to added but no more independent information. The four blocks show great consistency in terms of $DII$ \textit{vs.} number of non-zero features (Fig.  \ref{fig:DII4blocks}), meaning that they hold similar information for the same number of non-zero features.

\begin{figure}[!ht]
    \centering
    \includegraphics[width=.99\textwidth]{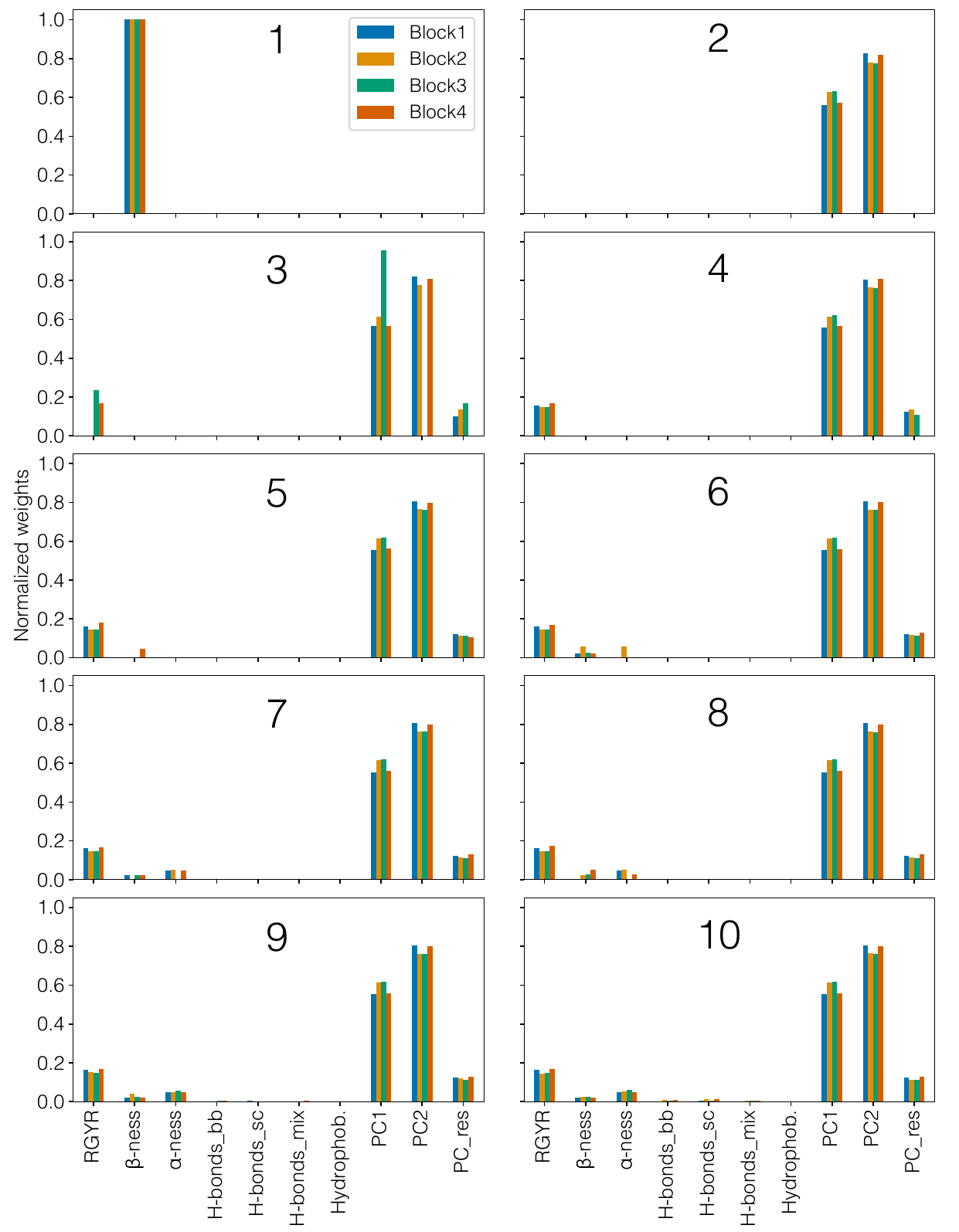}
    \caption{\textbf{Feature weights from block analysis.} Relative feature weights for each number of non-zero features (header in each plot) and each block, corresponding to Fig. \ref{fig:DII4blocks}. The ten input features are: Radius of gyration (RGYR); anti-$\beta$-sheet content ($\beta$-ness); $\alpha$-helical content ($\alpha$-ness); the number of hydrogen bonds in the backbone (H-bonds\_bb), in the side chains (H-bonds\_sc), and between the backbone and side chains (H-bonds\_mix); the number of hydrophobic contacts (Hydroph.), principal component 1 (PC1); principal component 2 (PC2) and the principal component residuals (PC\_res).
    Source data are provided as a Source Data file.}
    \label{fig:DII4blocks_features}
\end{figure}

The input variables on which we performed the feature selection were the same collective variables as previously (see suppl. section \hyperref[sec:CLNsupp1]{1} and for details on the CVs see Methods of the main paper). Fig. \ref{fig:DII4blocks_features} shows the normalized (to the unit vector) relative weights for the selected number of non-zero features and each block, corresponding to Fig. \ref{fig:DII4blocks}.

The first observation in Fig. \ref{fig:DII4blocks_features} is that the $DII$ is not a submodular function of the number of features, meaning that the most informative single-feature, which is the anti-$\beta$-sheet content, is not part of the most informative duplets, triplets or quadruplets of features, whose main features always contain the principal components, and never the anti-$\beta$-sheet content. Secondly, the chosen features are mostly in good agreement across the four trajectory blocks. In certain instances, such as the most informative feature triplets, there are two competing features (RGYR and the PC residuals) which complement PC1 and PC2 and lead to similar $DII$s. In the chosen sets of four features this conflict resolves and both of these features form part of very consistent quadruplets. Finally, in almost any feature tuple bigger than two features, PC2 is the most important feature - a non-trivial observation.

The analyses show that independent trajectory parts the CLN025 MD simulation lead to consistently chosen features with consistent values of the $DII$.



\newpage
\section{Choosing tuples by DII}
\begin{figure}[!ht]
    \centering
    \includegraphics[width=.75\textwidth]{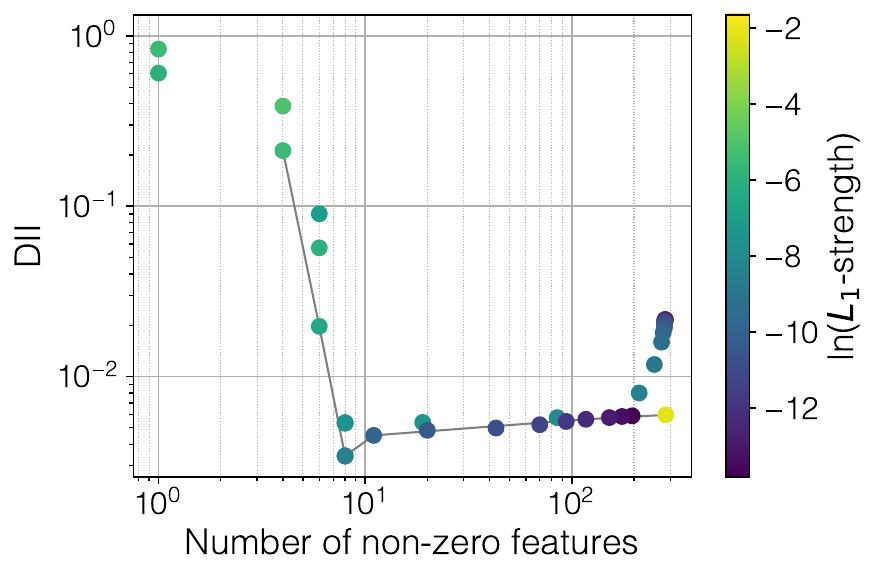}
    \caption{\textbf{$\boldsymbol{DII}$ results in L$_1$-search.} The resulting Differentiable Information Imbalances ($DII$s) for a L$_1$-search with many L$_1$ strengths plotted as a function of non-zero features on a log-log graph. Several different L$_1$ strengths can lead to the same number of non-zero features with different features and/or weights. The gray line corresponds to the $DII$ represented in Fig. 1B II of the main paper.
    Source data are provided as a Source Data file.}
    \label{fig:choosetuple}
\end{figure}

The example plotted in Fig \ref{fig:choosetuple} is a  L$_1$-search of the 285 monomials of the ten Gaussian random variables as input, with ten of them scaled as ground truth. The figure shows how several different L$_1$ strengths lead to the same number of non-zero features with different features and/or weights. In these cases, the lowest $DII$s per numbers of non-zero features should be selected.

\section{Pseudocodes}

We provide in Algorithm \ref{alg:pseudocode} a simple pseudocode for the optimization of the Differentiable Information Imbalance. The hyperparameters include the number of training epochs $n_{\text{epochs}}$, the starting learning rate $\eta_0$ (which is reduced with an exponential or cosine decay during the training, if \texttt{decaying\_lr} is "exp" or "cos"), the strength of the L$_1$ regularization $p$ and the softmax parameter $\lambda_0$. 
If $\lambda_0$ is not set by the user, the adaptive scheme is applied.
Additionally, if no initial value $\boldsymbol{w}_0$ of the scaling weights is set, the algorithm automatically sets it to the inverse of the features' standard deviations. Similarly, if no starting learning rate is provided, the algorithm chooses a suitable one. The value of the $\alpha$-component of the weight vector at epoch $t$ is denoted by $\boldsymbol{w}^\alpha_t$ ($\alpha=1,...,D)$.

In Algorithm \ref{alg:pseudocode2} we describe the backward greedy approach for the search of the optimal subsets of $D'$ features ($D' < D$). Here, the notation $\boldsymbol{w}^{(D')}$ is used to denote a weight vector with $D'$ non-zero components, and the standard deviation of feature $\alpha$ is written as std$^\alpha$. In each optimization, setting a component of the initial weight vector $\boldsymbol{w}_0$ to zero is equivalent to removing the corresponding feature from space $A$, as the derivative of the $DII$ with respect to $w^\alpha$ is equal to zero throughout the entire training by setting $w_0^\alpha = 0$ (see Eq. (6) in the main text).

\SetKwComment{Comment}{/* }{ */}

\begin{algorithm}
\caption{Pseudocode for the optimization of $DII\left(d^A(\boldsymbol{w})\rightarrow d^B\right)$.}\label{alg:pseudocode}
\textbf{Parameters:} $n_{\text{epochs}}$, $\eta_0$, $\boldsymbol{w}_0$, $\lambda_0$\\
Compute rank matrix in space $B$: $r_{ij}^B$\;
$t = 0$ \Comment*[r]{epoch index}
Compute and save starting $DII$: $DII_0 = DII\left( d^A(\boldsymbol{w}_0)\rightarrow d^B\right)$\;
\While{\textnormal{$t < n_{\text{epochs}}$}}{
  Compute $\boldsymbol{w}_t$-dependent distances in space $A$: $d_{ij}^A(\boldsymbol{w}_t)$\;
  Compute softmax coefficients: $c_{ij}(\lambda,d^A(\boldsymbol{w}_t))$\;
  \eIf{\textnormal{$\lambda_0$ \textbf{is} \texttt{None}}}{
    Compute $\lambda_t$ given the current distances: $\lambda_t = \lambda_t(d_{ij}^A(\boldsymbol{w}_t))$ \Comment*[r]{adaptive lambda}
  }{
    $\lambda_t = \lambda_0$\;
  }
  \eIf{\textnormal{\texttt{decaying\_lr} \textbf{is} \texttt{True}}}{
    Compute $\eta_t$ according to chosen schedule \Comment*[r]{set learning rate}
  }{
    $\eta_t = \eta_0$\;
  }
  Compute gradient of $DII$: $\nabla_{\boldsymbol{w}_t} DII\left( d^A(\boldsymbol{w}_t)\rightarrow d^B\right)$\;
  $\boldsymbol{w}_{t+1} = \boldsymbol{w}_t - \eta_t\, \nabla_{\boldsymbol{w}_t} DII\left( d^A(\boldsymbol{w}_t)\rightarrow d^B\right)$ \Comment*[r]{gradient descent step}
  \If{\textnormal{$p \neq 0$}}{
    \For{$\alpha=1,...,D$}{
        \eIf{$w^{\alpha}_{t+1} > 0$}{
            $w^{\alpha}_{t+1} = \max(0, w^{\alpha}_{t+1} - \eta_t p) $ \Comment*[r]{L$_1$ regularization step}
        }{
            $w^{\alpha}_{t+1} = |\min(0, w^{\alpha}_{t+1} + \eta_t p)| $\;
        }
    }
  }
  Compute and save $DII$ with new weights: $DII_{t+1} = DII\left( d^A(\boldsymbol{w}_{t+1})\rightarrow d^B\right)$\;
  $t = t + 1$\;
}
\KwResult{$DII_t,\, \boldsymbol{w}_t$ for $t = 0,...,n_{\text{epochs}}$}
\end{algorithm}

\begin{algorithm}
\caption{Pseudocode for the backward greedy optimization the $DII$.}\label{alg:pseudocode2}

$D' = D$ \Comment*[r]{number of non-zero components}
$\boldsymbol{w}_{0} = (1/\text{std}^1,\,1/\text{std}^2,...,1/\text{std}^D)$ \Comment*[r]{initialize starting weight vector}
\While{\textnormal{$D' > 0$}}{
    Optimize $DII(\boldsymbol{w}_{0})$ according to Alg. (1) and extract optimized weights $\hat{\boldsymbol{w}} = \arg\min_{\boldsymbol{w}} DII$; 
    
    Save the optimal $D'$ non-zero weights: $\hat{\boldsymbol{w}}^{(D')} = \hat{\boldsymbol{w}}$\;
    Set the smallest among the $D'$ non-zero weights to zero: $\min \hat{w} =0$\;
    Set the new initial weight vector: $\boldsymbol{w}_{0} = \hat{\boldsymbol{w}}$\;
    $D' = D' -1$ \Comment*[r]{reduce the dimensionality by 1}
}
\KwResult{$\hat{\boldsymbol{w}}^{(D')}$ for $D' = 1,...,D$}
\end{algorithm}

\clearpage
\newpage
\printbibliography